%% file: main.tex
\documentclass[acmsmall]{acmart}
\AtBeginDocument{%
  \providecommand\BibTeX{{%
    \normalfont B\kern-0.5em{\scshape i\kern-0.25em b}\kern-0.8em\TeX}}}

\setcopyright{acmcopyright}
\copyrightyear{2023}
\acmYear{2023}

\usepackage{subcaption}
\usepackage{booktabs,multirow}
\usepackage{enumitem}
\usepackage{graphicx}

\usepackage{amssymb}
\usepackage{pifont}
\usepackage{color, colortbl}

\definecolor{LightCyan}{rgb}{0.88,1,1}
\definecolor{Gray}{gray}{0.9}
\newcommand{\cmark}{\ding{51}}%
\newcommand{\xmark}{\ding{55}}%

\begin{document}

\title{Data-centric Artificial Intelligence: A Survey}

\author{Daochen Zha}
\email{daochen.zha@rice.edu}
\affiliation{%
  \institution{Rice University}
  \country{United States}
}

\author{Zaid Pervaiz Bhat}
\email{zaid.bhat1234@tamu.edu}
\affiliation{%
  \institution{Texas A\&M University}
  \country{United States}
}

\author{Kwei-Herng Lai}
\email{khlai@rice.edu}
\affiliation{%
  \institution{Rice University}
  \country{United States}
}

\author{Fan Yang}
\email{fyang@rice.edu}
\affiliation{%
  \institution{Rice University}
  \country{United States}
}

\author{Zhimeng Jiang}
\email{zhimengj@tamu.edu}
\affiliation{%
  \institution{Texas A\&M University}
  \country{United States}
}

\author{Shaochen Zhong}
\email{hz88@rice.edu}
\affiliation{%
  \institution{Rice University}
  \country{United States}
}

\author{Xia Hu}
\email{Xia.Hu@rice.edu}
\affiliation{%
  \institution{Rice University}
  \country{United States}
}



\begin{abstract}

Artificial Intelligence (AI) is making a profound impact in almost every domain. A vital enabler of its great success is the availability of abundant and high-quality data for building machine learning models. Recently, the role of data in AI has been significantly magnified, giving rise to the emerging concept of \emph{data-centric AI}. The attention of researchers and practitioners has gradually shifted from advancing model design to enhancing the quality and quantity of the data. In this survey, we discuss the necessity of data-centric AI, followed by a holistic view of three general data-centric goals (training data development, inference data development, and data maintenance) and the representative methods. We also organize the existing literature from automation and collaboration perspectives, discuss the challenges, and tabulate the benchmarks for various tasks. We believe this is the first comprehensive survey that provides a global view of a spectrum of tasks across various stages of the data lifecycle. We hope it can help the readers efficiently grasp a broad picture of this field, and equip them with the techniques and further research ideas to systematically engineer data for building AI systems. A companion list of data-centric AI resources will be regularly updated on \url{https://github.com/daochenzha/data-centric-AI}



\end{abstract}





\maketitle

\input{introduction}

\input{sec2}

\input{sec3}

\input{sec4}

\input{sec5}

\input{sec6}

\input{sec7}

\input{sec8}

\bibliographystyle{acm}
\bibliography{paper}

\end{document}

%% file: introduction.tex
\section{Introduction}

\begin{figure}[t]
  \centering
  \begin{subfigure}[b]{0.545\textwidth}
    \centering
    \includegraphics[width=1.0\textwidth]{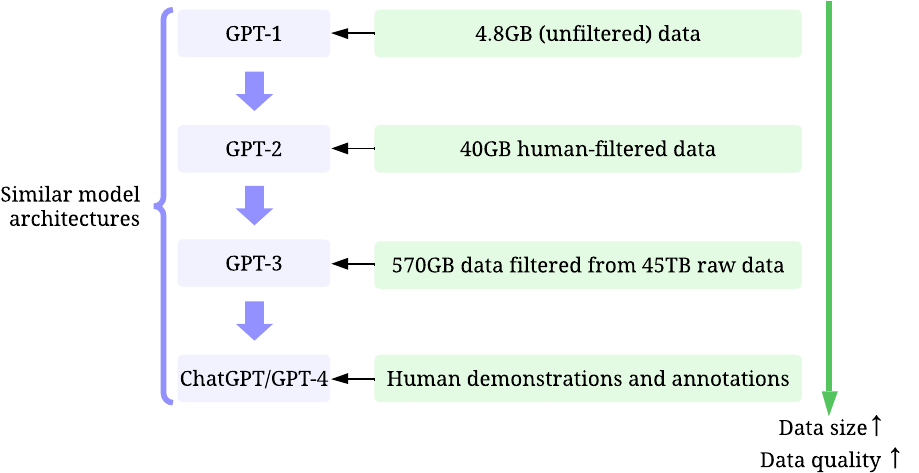}
  \end{subfigure}%
  \begin{subfigure}[b]{0.455\textwidth}
    \centering
    \includegraphics[width=1.0\textwidth]{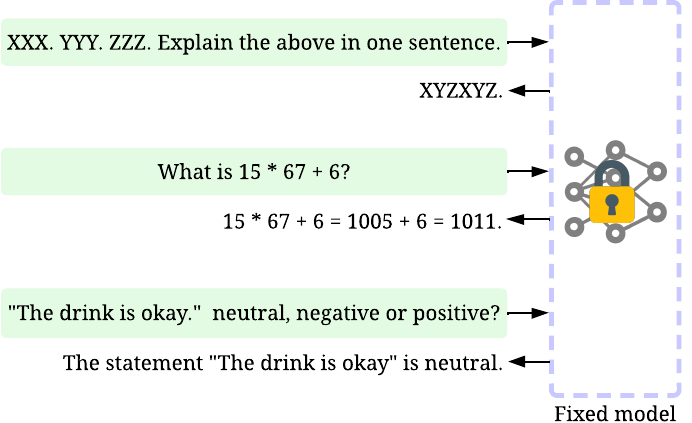}
  \end{subfigure}%
  \vspace{-7pt}
  \caption{Motivating examples that highlight the central role of data in AI. On the left, large and high-quality training data are the driving force of recent successes of GPT models, while model architectures remain similar, except for more model weights. The detailed data collection strategies of GPT models are provided in~\cite{zhu2015aligning,radford2018improving,radford2019language,brown2020language,ouyang2022training,gpt4}. On the right, when the model becomes sufficiently powerful, we only need to engineer prompts (inference data) to accomplish our objectives, with the model being fixed.}
  \vspace{-10pt}
  \label{fig:motivation}
\end{figure}

The past decade has witnessed dramatic progress in Artificial Intelligence (AI), which has made a profound impact in almost every domain, such as natural language processing~\cite{chowdhary2020natural}, computer vision~\cite{voulodimos2018deep}, recommender system~\cite{zhang2019deep}, healthcare~\cite{miotto2018deep}, biology~\cite{webb2018deep}, finance~\cite{ozbayoglu2020deep}, and so forth. A vital enabler of these great successes is the availability of abundant and high-quality data. Many major AI breakthroughs occur only after we have the access to the right training data. For example, AlexNet~\cite{krizhevsky2017imagenet}, one of the first successful convolutional neural networks, was designed based on the ImageNet dataset~\cite{deng2009imagenet}. AlphaFold~\cite{jumper2021highly}, a breakthrough of AI in scientific discovery, will not be possible without annotated protein sequences~\cite{mirdita2017uniclust}. The recent advances in large language models rely on large text data for training~\cite{kenton2019bert,radford2018improving,radford2019language,brown2020language} (left of Figure~\ref{fig:motivation}). Besides training data, well-designed inference data has facilitated the initial recognition of numerous critical issues in AI and unlocked new model capabilities. A famous example is adversarial samples~\cite{kurakin2018adversarial} that confuse neural networks through specialized modifications of input data, which causes a surge of interest in studying AI security.
Another example is prompt engineering~\cite{liu2023pre}, which accomplishes various tasks by solely tuning the input data to probe knowledge from the model while keeping the model fixed (right of Figure~\ref{fig:motivation}). In parallel, the value of data has been well-recognized in industries. Many big tech companies have built infrastructures to organize, understand, and debug data for building AI systems~\cite{thusoo2010data,barclay2000microsoft,armbrust2021lakehouse,varia2014overview}. All these efforts in constructing training data, inference data, and the infrastructure to maintain data have paved the path for the achievements in AI today.

Recently, the role of data in AI has been significantly magnified, giving rise to the emerging concept of \emph{data-centric AI}~\cite{polyzotis2021can,jarrahi2022principles,jakubik2022data,zha2023data,whang2023data}. In the conventional model-centric AI lifecycle, researchers and developers primarily focus on identifying more effective models to improve AI performance while keeping the data largely unchanged. However, this model-centric paradigm overlooks the potential quality issues and undesirable flaws of data, such as missing values, incorrect labels, and anomalies. Complementing the existing efforts in model advancement, data-centric AI emphasizes the systematic engineering of data to build AI systems, shifting our focus from model to data. It is important to note that ``data-centric'' differs fundamentally from ``data-driven'', as the latter only emphasizes the use of data to guide AI development, which typically still centers on developing models rather than engineering data.

Several initiatives have already been dedicated to the data-centric AI movement. A notable one is a competition launched by Ng et al.~\cite{ng2021data}, which asks the participants to iterate on the dataset only to improve the performance. Snorkel~\cite{ratner2017snorkel} builds a system that enables automatic data annotation with heuristic functions without hand labeling. A few rising AI companies have placed data in the central role because of many benefits, such as improved accuracy, faster deployment, and standardized workflow~\cite{landingai,snorkelai,scaleai}. These collective initiatives across academia and industry demonstrate the necessity of building AI systems using data-centric approaches.

With the growing need for data-centric AI, various methods have been proposed. Some relevant research subjects are not new. For instance, data augmentation~\cite{feng2021survey} has been extensively investigated to improve data diversity. Feature selection~\cite{li2017feature} has been studied since decades ago for preparing more concise data. Meanwhile, some new research directions have emerged recently, such as data programming for labeling data quickly~\cite{ratner2016data}, algorithmic
recourse for understanding model decisions~\cite{karimi2021algorithmic}, and prompt engineering that modifies the input of large language models to obtain the desirable predictions~\cite{liu2023pre}. From another dimension, some works are dedicated to making data processing more automated, such as automated data augmentation~\cite{cubuk2019autoaugment}, and automated pipeline discovery~\cite{drori2021alphad3m,lai2021tods}. Some other methods emphasize human-machine collaboration in creating data so that the model can align with human intentions. For example, the remarkable success of ChatGPT and GPT-4~\cite{gpt4} is largely attributed to the reinforcement learning from human feedback procedure~\cite{christiano2017deep}, which asks humans to provide appropriate responses to prompts and rank the outputs to serve as the rewards~\cite{ouyang2022training}. Although the above methods are independently developed for different purposes, their common objective is to ensure data quality, quantity, and reliability so that the models behave as intended.

Motivated by the need for data-centric AI and the numerous proposed methods, this survey provides a holistic view of the technological advances in data-centric AI and summarizes the existing research directions. In particular, this survey centers on the following research questions:



\begin{itemize}
    \item RQ1: What are the necessary tasks to make AI data-centric?
    \item RQ2: Why is automation significant for developing and maintaining data?
    \item RQ3: In which cases and why is human participation essential in data-centric AI?
    \item RQ4: What is the current progress of data-centric AI?
\end{itemize}

By answering these questions, we make three contributions. \emph{Firstly}, we provide a comprehensive overview to help readers efficiently grasp a broad picture of data-centric AI from different perspectives, including definitions, tasks, algorithms, challenges, and benchmarks. \emph{Secondly}, we organize the existing literature under a goal-driven taxonomy. We further identify whether human involvement is needed in each method and label the method with a level of automation or a degree of human participation. \emph{Lastly}, we analyze the existing research and discuss potential future opportunities.

This survey is structured as follows. Section~\ref{sec:2} presents an overview of the concepts and tasks related to data-centric AI. Then, we elaborate on the needs, representative methods, and challenges of three general data-centric AI goals, including training data development (Section~\ref{sec:3}), inference data development (Section~\ref{sec:4}), and data maintenance (Section~\ref{sec:5}). Section~\ref{sec:6} summarizes benchmarks for various tasks. Section~\ref{sec:7} discusses data-centric AI from a global view and highlights the potential future directions. Finally, we conclude this survey in Section~\ref{sec:8}.

%% file: sec2.tex
\section{Background of Data-centric AI}
\label{sec:2}

This section provides a background of data-centric AI. Section~\ref{sec:2:1} defines the relevant concepts. Section~\ref{sec:2:2} discusses why data-centric AI is needed. Section~\ref{sec:2:3} draws a big picture of the related tasks and presents a goal-driven taxonomy to organize the existing literature. Section~\ref{sec:2:4} focuses on automation and human participation in data-centric AI.

\subsection{Definitions}
\label{sec:2:1}

Researchers have described data-centric AI in different ways. Ng et al. defined it as ``the discipline of systematically engineering the data used to build an AI system''~\cite{datacentricaihub}. Polyzotis and Zaharia described it as ``an exciting new research field that studies the problem of constructing high-quality datasets for machine learning''~\cite{polyzotis2021can}. Jarrahi et al. mentioned that data-centric AI ``advocates for a systematic and iterative approach to dealing with data issues''~\cite{jarrahi2022principles}. Miranda noted that data-centric AI focuses on the problems that ``do not only involve the type of model to use, but also the quality of data at hand''~\cite{miranda2021towards}. While all these descriptions have emphasized the importance of data, the scope of data-centric AI remains ambiguous, i.e., what tasks and techniques belong to data-centric AI. Such ambiguity could prevent us from grasping a concrete picture of this field. Before starting the survey, it is essential to define some relevant concepts:

\begin{itemize}
    \item \textbf{Artificial Intelligence (AI)}: AI is a broad and interdisciplinary field that tries to enable computers to have human intelligence to solve complex tasks~\cite{winston1984artificial}. A dominant technique for AI is machine learning, which leverages data to train predictive models to accomplish some tasks.
    \item \textbf{Data}: Data is a very general concept to describe a collection of values that convey information. In the context of AI, data is used to train machine learning models or serve as the model input to make predictions. Data can appear in various formats, such as tabular data, images, texts, audio, and video.
    \item \textbf{Training Data}: Training data is the data used in the training phase of machine learning models. The model leverages training data to adjust its parameters and make predictions.
    \item \textbf{Inference Data}: Inference data is the data used in the inference phase of machine learning models. On the one hand, it can evaluate the performance of the model after it has been trained. On the other hand, tuning the inference data can help obtain the desirable outputs, such as tuning prompts for language models~\cite{liu2023pre}.
    \item \textbf{Data Maintenance}: Data maintenance refers to the process of maintaining the quality and reliability of data, which often involves efficient algorithms, tools, and infrastructures to understand and debug data. Data maintenance plays a crucial role in AI since it ensures training and inference data are accurate and consistent~\cite{jain2020overview}.
    \item \textbf{Data-centric AI}: Data-centric AI refers to a framework to develop, iterate, and maintain data for AI systems~\cite{zha2023data}. Data-centric AI involves the tasks and methods for building effective training data, designing proper inference data, and maintaining the data.
\end{itemize}

\subsection{Need for Data-centric AI} 
\label{sec:2:2}

In the past, AI was often viewed as a model-centric field, where the focus was on advancing model designs given fixed datasets.
However, the overwhelming reliance on fixed datasets does not necessarily lead to better model behavior in real-world applications, as it overlooks the breadth, difficulty, and fidelity of data to the underlying problem~\cite{mazumder2022dataperf}. Moreover, the models are often difficult to transfer from one problem to another since they are highly specialized and tailored to specific problems. Furthermore, undervaluing data quality could trigger data cascades~\cite{sambasivan2021everyone}, causing negative effects such as decreased accuracy and persistent biases~\cite{buolamwini2018gender}. This can severely hinder the applicability of AI systems, particularly in high-stakes domains.
 
Consequently, the attention of researchers and practitioners has gradually shifted toward data-centric AI to pursue data excellence~\cite{aroyo2022data}. 
Data-centric AI places a greater emphasis on enhancing the quality and quantity of the data with the model relatively more fixed. While this transition is still ongoing, we have already witnessed several accomplishments that shed light on its benefits. For example, the advancement of large language models is greatly dependent on the use of huge datasets~\cite{kenton2019bert,radford2018improving,radford2019language,brown2020language}. Compared to GPT-2~\cite{radford2019language}, GPT-3~\cite{brown2020language} only made minor modifications in the neural architecture while spending efforts collecting a significantly larger high-quality dataset for training. ChatGPT~\cite{ouyang2022training}, a remarkably successful application of GPT-3, adopts a similar neural architecture as GPT-3 and uses a reinforcement learning from human feedback procedure~\cite{christiano2017deep} to generate high-quality labeled data for fine-tuning. A new approach, known as prompt engineering~\cite{liu2023pre}, has seen significant success by focusing solely on tuning data inputs. The benefits of data-centric approaches can also be validated by practitioners~\cite{landingai,snorkelai,scaleai}. For instance, Landing AI, a computer vision company, observes improved accuracy, reduced development time, and more consistent and scalable methods from the adoption of data-centric approaches~\cite{landingai}. All these achievements demonstrate the promise of data-centric AI.

It is noteworthy that data-centric AI does not diminish the value of model-centric AI. Instead, these two paradigms are complementarily interwoven in building AI systems. On the one hand, model-centric methods can be used to achieve data-centric AI goals. For example, we can utilize a generation model, such as GAN~\cite{goodfellow2020generative,zhang2019self} and diffusion model~\cite{ho2020denoising,kingma2021variational,rombach2022high}, to perform data augmentation and generate more high-quality data. On the other hand, data-centric AI could facilitate the improvement of model-centric AI objectives. For instance, the increased availability of augmented data could inspire further advancements in model design. Therefore, in production scenarios, data and models tend to evolve alternatively in a constantly changing environment~\cite{polyzotis2021can}.

\begin{figure}[t]
  \centering
    \includegraphics[width=0.85\textwidth]{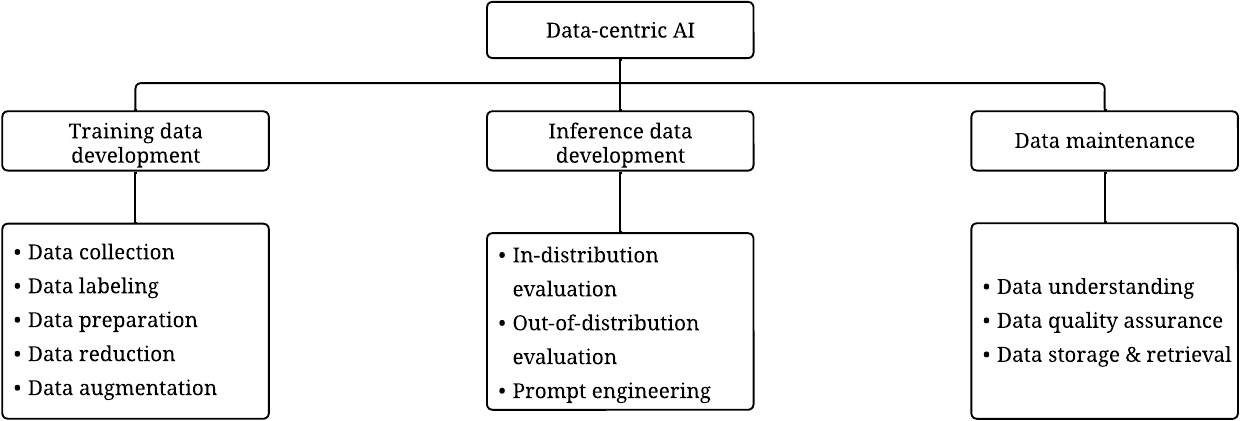}
  \caption{Data-centric AI framework.}
  \label{fig:taxonomy}
\end{figure}

\subsection{Tasks in Data-centric AI} 
\label{sec:2:3}

The ambitious movement to data-centric AI can not be achieved without making progress on concrete and specific tasks. Unfortunately, most of the existing literature has been focused on discussing the foundations and perspectives of data-centric AI without clearly specifying the associated tasks~\cite{polyzotis2021can,jarrahi2022principles,jakubik2022data,seedat2022dc}. As an effort to resolve this ambiguity, the recently proposed DataPerf benchmark~\cite{mazumder2022dataperf} has defined six data-centric AI tasks: training set creation, test set creation, selection algorithm, debugging algorithm, slicing algorithm, and valuation algorithm. However, this flat taxonomy can only partially cover the existing data-centric AI literature. For example, some crucial tasks such as data labeling~\cite{zhang2022survey} are not included. The selection algorithm only addresses instance selection but not feature selection~\cite{li2017feature}. The test set creation is restricted to selecting items from a supplemental set rather than generating a new set~\cite{saporta2002data}. Thus, a more nuanced taxonomy is necessary to fully encompass data-centric AI literature.

To gain a more comprehensive understanding of data-centric AI, we draw a big picture of the related tasks and present a goal-driven taxonomy to organize the existing literature in Figure~\ref{fig:taxonomy}. We divide data-centric AI into three goals: training data development, inference data development, and data maintenance, where each goal is associated with several sub-goals, and each task belongs to a sug-goal. We give a high-level overview of these goals below.


\begin{itemize}
    \item \textbf{Training data development}: The goal of training data development is to collect and produce rich and high-quality training data to support the training of machine learning models. It consists of five sub-goals, including 1) data collection for gathering raw training data, 2) data labeling for adding informative labels, 3) data preparation for cleaning and transforming data, 4) data reduction for decreasing data size with potentially improved performance, and 5) data augmentation for enhancing data diversity without collecting more data.

    \item \textbf{Inference data development}: The objective is to create novel evaluation sets that can provide more granular insights into the model or trigger a specific capability of the model with engineered data inputs.
    There are three sub-goals in this effort: 1) in-distribution evaluation and 2) out-of-distribution evaluation aim to generate samples that adhere to or differ from the training data distribution, respectively, while 3) prompt engineering tunes the prompt in language models to get the desired predictions. The tasks in inference data development are relatively open-ended since they are often designed to assess or unlock various capabilities of the model.
    

    \item \textbf{Data maintenance}: In real-world applications, data is not created once but rather necessitates continuous maintenance. The purpose of data maintenance is to ensure the quality and reliability of data in a dynamic environment. It involves three essential sub-goals: 1) data understanding, which targets providing visualization and valuation of the complex data, enabling humans to gain valuable insights, 2) data quality assurance, which develops quantitative measurements and quality improvement strategies to monitor and repair data, and 3) data storage \& retrieval, which aims to devise efficient algorithms to supply the data in need via properly allocating resources and efficiently processing queries. Data maintenance plays a fundamental and supportive role in the data-centric AI framework, ensuring that the data in training and inference is accurate and reliable.
\end{itemize}

Following the three general goals, we survey various data-centric AI tasks, summarized in Table~\ref{tbl:tasksummary}.

\begin{table}[t]
\centering
\caption{Representative tasks under the data-centric AI framework.}
\footnotesize	
\label{tbl:tasksummary}
\setlength{\tabcolsep}{0.2pt}
\begin{tabular}{l|l|l} \toprule
\textbf{Goal} & \textbf{Sub-goal} & \textbf{Tasks} \\
\midrule

\multirow{6}{*}{\shortstack[l]{Training data\\ development}} & \cellcolor{gray!10}Collection & \cellcolor{gray!10}Dataset discovery~\cite{bogatu2020dataset}, data integration~\cite{stonebraker2018data}, raw data synthesis~\cite{lai2021revisiting}\\

~ & ~ & Crowdsourced labeling~\cite{kutlu2020annotator}, semi-supervised labeling~\cite{zoph2020rethinking}, active learning~\cite{ren2021survey}, \\

~ & \multirow{-2}{*}{Labeling} & data programming~\cite{ratner2016data}, distant supervision~\cite{mintz2009distant} \\

~ & \cellcolor{gray!10}Preparation & \cellcolor{gray!10}Data cleaning~\cite{zhang2016missing}, feature extraction~\cite{salau2019feature}, feature transformation~\cite{ali2014data}  \\

~ & Reduction & Feature selection~\cite{li2017feature}, dimensinality reduction~\cite{abdi2010principal}, instance selection~\cite{riquelme2003finding} \\

~ & \cellcolor{gray!10}Augmentation &  \cellcolor{gray!10}Basic manipulation~\cite{zhang2018mixup}, augmentation data synthesis~\cite{frid2018synthetic}, upsampling~\cite{zha2022towards}\\

\midrule

\multirow{3}{*}{\shortstack[l]{Inference data\\ development}} & In-distribution & Data slicing~\cite{chung2019slice}, algorithmic recourse~\cite{karimi2021algorithmic}\\
~ & \cellcolor{gray!10}Out-of-distribution & \cellcolor{gray!10}Generating adversarial samples~\cite{moosavi2016deepfool}, generating samples with distribution shift~\cite{koh2021wilds}\\
~ & Prompt engineering & Manual prompt engineering~\cite{schick2020few}, automated prompt engineering~\cite{wallace2019universal} \\

\midrule

\multirow{4}{*}{\shortstack[l]{Data\\ maintenance}} & \cellcolor{gray!10}~ & \cellcolor{gray!10}Visual summarization~\cite{burch2014benefits}, clustering for visualization~\cite{fahad2014survey}, \\
~ & \cellcolor{gray!10}\multirow{-2}{*}{Understanding} & \cellcolor{gray!10}visualization recommendation~\cite{wongsuphasawat2015voyager}, valuation~\cite{ghorbani2020distributional} \\
~ & Quality assurance & Quality assessment~\cite{sadiq2018data}, quality improvement~\cite{baylor2017tfx}\\
~ & \cellcolor{gray!10}Storage \& retrieval & \cellcolor{gray!10}Resource allocation~\cite{herodotou2011starfish}, query index selection~\cite{sun2019end}, query rewriting~\cite{baik2019bridging} \\

\bottomrule
\end{tabular}
\end{table}

\subsection{Automation and Human Participation in Data-centric AI} 

\label{sec:2:4}


Data-centric AI consists of a spectrum of tasks related to different data lifecycle stages. To keep pace with the ever-growing size of the available data, in some data-centric AI tasks, it is imperative to develop automated algorithms to streamline the process. For example, there is an increasing interest in automation in data augmentation~\cite{cubuk2019autoaugment,zha2022towards}, and feature transformation~\cite{khurana2018feature}. Automation in these tasks will improve not only efficiency but also accuracy~\cite{mazumder2022dataperf}. Moreover, automation can facilitate the consistency of the results, reducing the chance of human errors. Whereas for some other tasks, human involvement is essential to ensure the data is consistent with our intentions. For example, humans often play an indispensable role in labeling data~\cite{zhang2022survey}, which helps machine learning algorithms learn to make the desired predictions. Whether human participation is needed depends on whether our objective is to align data with human expectations. In this survey, we categorize each paper into \emph{automation} and \emph{collaboration}, where the former focuses on automating the process, and the latter concerns human participation. Automation-oriented methods usually have different automation objectives. We can identify several levels of automation from the existing methods:


\begin{itemize}
    \item \textbf{Programmatic automation:} Using programs to deal with the data automatically. The programs are often designed based on some heuristics and statistical information.
    \item \textbf{Learning-based automation:} Learning automation strategies with optimization, e.g., minimizing an objective function. The methods at this level are often more flexible and adaptive but require additional costs for learning.
    \item \textbf{Pipeline automation:} Integrating and tuning a series of strategies across multiple tasks, which could help identify globally optimal strategies. However, tuning may incur significantly more costs.
\end{itemize}





    


Note that this categorization does not intend to differentiate good and bad methods. For example, a pipeline automation method may not necessarily be better than programmatic automation solutions since it could be over-complicated in many scenarios. Instead, we aim to show insight into how automation has been applied to different data-centric goals and understand the literature from a global view. From another perspective, collaboration-oriented methods often require human participation in different forms. We can identify several degrees of human participation:

\begin{figure}[t]
  \centering
    \includegraphics[width=0.85\textwidth]{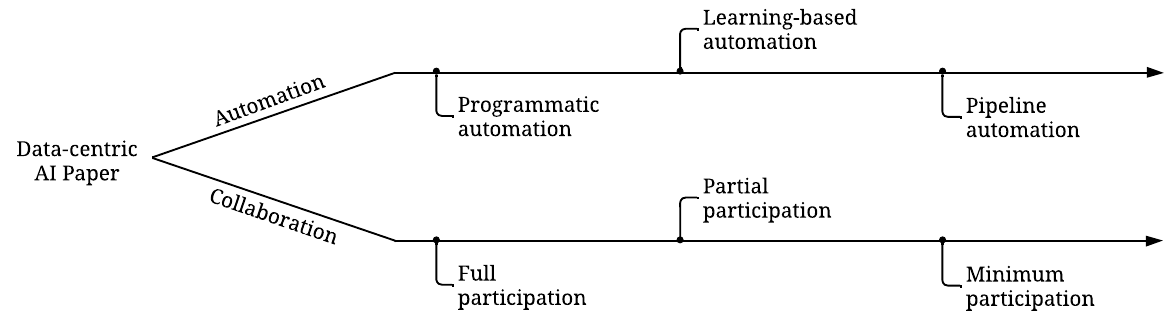}
  \caption{Data-centric AI papers are categorized into automation and collaboration depending on whether human participation is needed. Each method has a different level of automation or requires a different degree of human participation.}
  \label{fig:surveystrategy}
\end{figure}

\begin{itemize}
    \item \textbf{Full participation:} Humans fully control the process. The method assists humans in making decisions. The methods that require full participation can often align well with human intentions but can be costly.
    \item \textbf{Partial participation:} The method is in control of the process. However, humans need to intensively or continuously supply information, e.g., by providing a large amount of feedback or frequent interactions.
    \item \textbf{Minimum participation:} The method is in full control of the whole process and only consults humans when needed. Humans only participate when prompted or asked to do so. The methods that belong to this degree are often more desirable when encountering a massive amount of data and a limited budget for human efforts.
\end{itemize}

Similarly, the degree of human participation, to a certain extent, only reflects the tradeoff between efficiency (less human labor) and effectiveness (better aligned with humans). The selection of methods depends on the application domain and stakeholders' needs. To summarize, we design Figure~\ref{fig:surveystrategy} to organize the existing data-centric AI papers. We assign each paper to either a level of automation or a degree of human participation.

Some previous surveys only focus on specific scopes of data-centric AI, such as data augmentation~\cite{feng2021survey,wen2021time,shorten2019survey}, data labeling~\cite{zhang2022survey}, and feature selection~\cite{li2017feature}. The novelty of our paper is that it provides a holistic view of the tasks, methods, and benchmarks by providing a goal-driven taxonomy to organize the tasks followed by an automation- and collaboration-oriented design to categorize methods. Moreover, we discuss the needs, challenges, and future directions from the broad data-centric AI view, aiming to motivate collective initiatives to push forward this field.





%% file: sec3.tex
\begin{figure}[t]
  \centering
    \includegraphics[width=\textwidth]{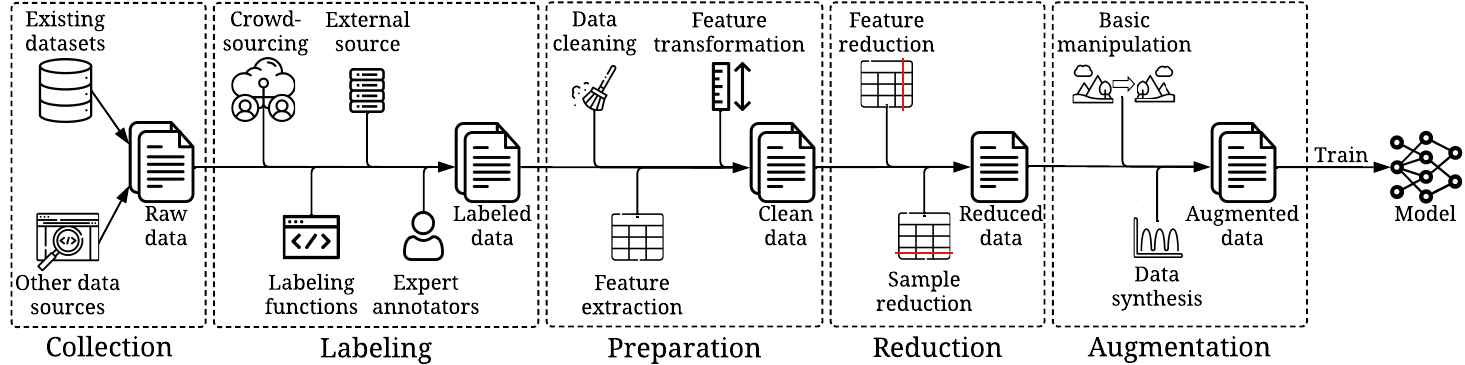}
  \caption{An overview of training data development. Note that the figure illustrates only a general pipeline, and not all steps are mandatory. For instance, unsupervised learning does not require data labeling. These steps can be executed in a different order as well. For example, data augmentation can occur before data reduction.}
  \label{fig:trainingdevelopmentoverview}
\end{figure}

\section{Training Data Development}
\label{sec:3} 

Training data provides the foundation for machine learning models, as the model performance is heavily influenced by its quality and quantity. In this section, we summarize the essential steps to create and process training data, visualized in Figure~\ref{fig:trainingdevelopmentoverview}. Data creation focuses on effectively and efficiently encoding human intentions into datasets, including data collection~(Section~\ref{sec:3:1}) and data labeling~(Section~\ref{sec:3:2}). Data processing aims to make data suitable for learning, including data preparation~(Section~\ref{sec:3:3}), data reduction~(Section~\ref{sec:3:4}), and data augmentation~(Section~\ref{sec:3:5}). After introducing these steps, we discuss pipeline search~(Section~\ref{sec:3:6}), an emerging trend that aims to connect them and search for the most effective end-to-end solution. Table~\ref{tbl:trainingmethodsummary} summarizes the representative tasks and methods for training data development.

\subsection{Data Collection}
\label{sec:3:1}

Data collection is the process of gathering and acquiring data from various sources, which fundamentally determines data quality and quantity. This process heavily relies on domain knowledge. With the increasing availability of data, there has been a surge in the development of efficient strategies to leverage existing datasets. In the following, we discuss the role of domain knowledge, an overview of more efficient data collection strategies, and challenges.

\subsubsection{Role of Domain Knowledge}
A deep understanding of the application domain or industry is critical for collecting relevant and representative data. For example, when building a recommendation system, it is crucial to decide what user/item features to collect based on the application domain~\cite{zhang2019deep}. The domain-specific knowledge can also help in synthesizing data. For instance, knowledge about financial markets and trading strategies can facilitate the generation of more realistic synthetic anomalies~\cite{lai2021revisiting}. Domain knowledge is essential for effective data collection since it helps align data with the intentions of stakeholders and ensure the data is relevant and representative.

\subsubsection{Efficient Data Collection Strategies}
Traditionally, datasets are constructed from scratch by manually collecting the relevant information. However, this process is time-consuming. More efficient methods have been developed by leveraging the existing data. Here, we describe the methods for dataset discovery, data integration, and data synthesis.

\emph{Dataset discovery.} As the number of available datasets continuously grows, it becomes possible to amass the existing datasets of interest to construct a new dataset that meets our needs. Given a human-specified query (e.g., the expected attribute names), dataset discovery aims to identify the most related and useful datasets from a data lake, a repository of datasets stored in
its raw formats, such as public data-sharing platforms~\cite{bhardwaj2015datahub} and data marketplaces. The existing research for dataset discovery mainly differs in calculating relatedness. A representative strategy is to abstract the datasets as a graph, where the nodes are columns of the data sources, and edges represent relationships between two nodes~\cite{fernandez2018aurum}. Then a tailored query language is designed to allow users to express complex query logic to retrieve the relevant datasets. Another approach is table union search~\cite{nargesian2018table}, which measures the unionability of datasets based on the overlapping of the attribute values. Recent work measures the relatedness in a more comprehensive way by considering attribute names, value overlapping, word embedding, formats, and domain distributions~\cite{bogatu2020dataset}. All these methods can significantly reduce human labor in dataset discovery, as humans only need to provide queries.

\emph{Data integration.} Given a few datasets from different sources, data integration aims to combine them into a unified dataset. The difficulty lies in matching the columns across datasets and transforming the values of data records from the source dataset to the target dataset. Traditional solutions rely on rule-based systems~\cite{lenzerini2002data,kumar2016join}, which can not scale. Recently, machine learning has been utilized to automate the data integration process in a more scalable way~\cite{stonebraker2013data,stonebraker2018data}. For example, the transformation of data values can be formulated as a classification problem, where the input is the data value from the source dataset, and the output is the transformed value from the target dataset~\cite{stonebraker2018data}. Then we can train a classifier with the training data generated by rules and generalize it to unseen data records. The automated data integration techniques make it possible to merge a larger number of existing datasets efficiently.

\emph{Raw data synthesis.} In some scenarios, it is more efficient to synthesize a dataset that contains the desirable patterns than to collect these patterns from the real world. A typical scenario is anomaly detection, where it is often hard to collect sufficient real anomalies since they can be extremely rare. Thus, researchers often insert anomaly patterns into anomaly-free datasets. For example, a general anomaly synthesis criterion has been proposed for time series data~\cite{lai2021revisiting}, where a time series is modeled as a parameterized combination of trend, seasonality, and shapelets. Then different point- and pattern-wise anomalies can be generated by altering these parameters. However, such synthesis strategies may not be suitable for all domains. For example, the anomaly patterns in financial time series can be quite different from those from electricity time series. Thus, properly designing data synthesis strategies still requires domain knowledge.


\subsubsection{Challenges}
Data collection is a very challenging process that requires careful planning. From the technical perspective, datasets are often diverse and not well-aligned with each other, so it is non-trivial to measure their relatedness or integrate them appropriately. Effectively synthesizing data from the existing dataset is also tricky, as it heavily relies on domain knowledge. Moreover, some critical issues during data collection can not be resolved solely from a technical perspective. For example, in many real-world situations, we may be unable to locate a readily available dataset that aligns with our requirements so we still have to collect data from the ground up. However, some data sources can be difficult to obtain due to legal, ethical, or logistical reasons. Collecting new data also involves ethical considerations, particularly with regard to informed consent, data privacy, and data security. Researchers and practitioners must be aware of these challenges in studying and executing data collection.

\begin{table}[t]
\footnotesize
\centering
\setlength{\tabcolsep}{3.5pt}
\caption{Papers for achieving different sub-goals of training data development.}
\label{tbl:trainingmethodsummary}
\begin{tabular}{l|l|l|l|l} \toprule
\multirow{2}{*}{\textbf{Sub-goal}} & \multirow{2}{*}{\textbf{Task}} & \multirow{2}{*}{\textbf{Method type}} &  \multirow{2}{*}{\textbf{\shortstack[l]{Automation level/\\ participation degree }}} & \multirow{2}{*}{\textbf{Reference}} \\
~ & ~ & ~ & ~ & ~ \\
\midrule

\multirow{4}{*}{Collection} & Dataset discovery & Collaboration  &  Minimum & \cite{fernandez2018aurum,nargesian2018table,bogatu2020dataset} \\
~ & Data integration & Automation & Programmatic &  \cite{lenzerini2002data,kumar2016join} \\
~ & Data integration & Automation & Learning-based & \cite{stonebraker2013data,stonebraker2018data} \\
~ & Raw data synthesis & Automation & Programmatic &  \cite{lai2021revisiting} \\
\cline{1-5}

\multirow{6}{*}{Labeling} & Crowdsourced labeling & Collaboration & Full & \cite{kutlu2020annotator,tang2011semi,dekel2009vox} \\
~ & Semi-supervised labeling & Collaboration & Partial & \cite{zoph2020rethinking,zhou2004democratic,chong2020graph,ouyang2022training} \\
~ & Active learning & Collaboration &  Partial & \cite{cohn1996active,ren2021survey,dong2023active,zha2020meta} \\
~ & Data programming & Collaboration &  Partial & \cite{boeckinginteractive,galhotra2021adaptive} \\
~ & Data programming & Collaboration &  Minimum & \cite{ratner2016data,ratner2017snorkel,zha2019multi,hooper2021cut} \\
~ & Distant supervision & Automation &  Learning-based & \cite{mintz2009distant} \\
\cline{1-5}

\multirow{7}{*}{Preparation} & Data cleaning & Automation & Programmatic & \cite{zhang2016missing} \\
~ & Data cleaning & Automation & Learning-based & \cite{lakshminarayan1996imputation,heise2014estimating,jiang2022information,krishnan2019alphaclean} \\
~ & Data cleaning & Collaboration & Partial & \cite{wang2012crowder} \\
~ & Feature extraction & Automation & Programmatic & \cite{salau2019feature,barandas2020tsfel} \\
~ & Feature extraction & Automation & Learning-based & \cite{krizhevsky2017imagenet,wang2017time} \\
~ & Feature transformation & Automation & Programmatic & \cite{ali2014data,bisong2019introduction} \\
~ & Feature transformation & Automation & Learning-based & \cite{khurana2018feature} \\
\cline{1-5}

\multirow{6}{*}{Reduction} & Feature selection & Automation & Programmatic & \cite{thaseen2017intrusion,azhagusundari2013feature} \\
~ & Feature selection & Automation & Learning-based & \cite{yan2015feature,wang2015embedded}  \\
~ & Feature selection & Collaboration & Partial & \cite{zhang2019active,schnapp2021active}  \\
~ & Dimensionality reduction & Automation & Learning-based & \cite{abdi2010principal,xanthopoulos2013linear,bank2020autoencoders} \\
~ & Instance selection & Automation & Programmatic & \cite{riquelme2003finding,prusa2015using}\cite{liu2020mesa} \\
~ & Instance selection & Automation & Learning-based & \cite{sutton2012introduction,liu2020mesa} \\
\cline{1-5}

\multirow{5}{*}{Augmentation} & Basic manipulation & Automation & Programmatic & \cite{zhang2015character,zhang2018mixup,zhang2018mixup,wen2021time,chen2020mixtext,han2022g}. \\
~ & Basic manipulation & Automation & Learning-based & \cite{cubuk2019autoaugment} \\
~ & Augmentation data synthesis & Automation & Learning-based & \cite{frid2018synthetic,shorten2021text,hsu2017unsupervised,ho2022cascaded} \\
~ & Upsampling & Automation & Programmatic & \cite{chawla2002smote,he2008adasyn} \\
~ & Upsampling & Automation & Learning-based & \cite{zha2022towards} \\
\cline{1-5}

- & Pipeline search & Automation & Pipeline & \cite{feurer2015efficient,milutinovic2020evaluation,drori2021alphad3m,lai2021tods,zha2021autovideo,heffetz2020deepline,martinez2023towards} \\

\bottomrule
\end{tabular}
\end{table}

\subsection{Data Labeling}
\label{sec:3:2}

Data labeling is the process of assigning one or more descriptive tags or labels to a dataset, enabling algorithms to learn from and make predictions on the labeled data. Traditionally, this  is a time-consuming and resource-intensive manual process, particularly for large datasets. Recently, more efficient labeling methods have been proposed to reduce human efforts. In what follows, we discuss the need for data labeling, efficient labeling strategies, and challenges.

\subsubsection{Need for Data Labeling.}
Labeling plays a crucial role in ensuring that the model trained on the data accurately reflects human intentions. Without proper labeling, a model may not be able to make the desired predictions since the model can, at most, be as good as the data fed into it. Although unsupervised learning techniques are successful in domains such as large language models~~\cite{kenton2019bert,radford2018improving,radford2019language,brown2020language} and anomaly detection~\cite{pang2021deep}, the trained models may not well align with human expectations. Thus, to achieve a better performance, we often still need to fine-tune the large language models with human labels, such as ChatGPT~\cite{ouyang2022training}, and tune anomaly detectors with a small amount of labeled data~\cite{li2020pyodds,li2021automated,li2021autood,jiang2023weakly}. Thus, labeling data is essential for teaching models to align with and behave like humans.

\subsubsection{Efficient Labeling Strategies.}
Researchers have long recognized the importance of data labeling. Various strategies have been proposed to enhance labeling efficiency. We will discuss crowdsourced labeling, semi-supervised labeling, active learning, data programming, and distant supervision. Note that it is possible to combine them as hybrid strategies.

\emph{Crowdsourced labeling.}
Crowdsourcing is a classic approach that breaks down a labeling task into smaller and more manageable parts so that they can be outsourced and distributed to a large number of non-expert annotators. Traditional methods often only provide initial guidelines to annotators~\cite{yuen2011survey}. However, the guidelines can be unclear and ambiguous, so each annotator could judge the same situation subjectively and differently. One way to mitigate this inconsistency is to start with small pilot studies and iteratively refine the design of the labeling task~\cite{kutlu2020annotator}. Another is to ask multiple workers to annotate the same sample and infer a consensus label~\cite{tang2011semi}. Other studies focus on algorithmically improving label quality, e.g., pruning low-quality teachers~\cite{dekel2009vox}. All these crowdsourcing methods require full human participation but assist humans or enhance label quality in different ways. 

\emph{Semi-supervised labeling.} 
The key idea is to leverage a small amount of labeled data to infer the labels of the unlabeled data. A popular approach is self-training~\cite{zoph2020rethinking}, which trains a classifier based on labeled data and uses it to generate pseudo labels. To improve the quality of pseudo labels, a common strategy is to train multiple classifiers and find a consensus label, such as using different machine learning algorithms to train models on the same data~\cite{zhou2004democratic}. In parallel, researchers have studied graph-based semi-supervised labeling techniques~\cite{chong2020graph}. The idea is to construct a graph, where each node is a sample, and each edge represents the distance between the two nodes it connects. Then they infer labels through label propagation in the graph. Recently, a reinforcement learning from human feedback procedure is proposed~\cite{christiano2017deep} and used in ChatGPT~\cite{ouyang2022training}. They train a reward model based on human-labeled data and infer the reward for unlabeled data to fine-tune the language model. These semi-supervised labeling methods only require partial human participation to provide the initial labels.

\emph{Active learning.}
Active learning is an iterative labeling procedure that involves humans in the loop. In each iteration, the algorithm selects an unlabeled sample or batch of samples as a query for human annotation. The newly labeled samples help the algorithm choose the next query. The existing work mainly differs in query selection strategies. Early methods use statistical methods to estimate sample uncertainty and select the unlabeled sample the model is most uncertain about~\cite{cohn1996active}. Recent studies have investigated deep active learning, which leverages model output or designs specialized architectures to measure uncertainty~\cite{ren2021survey}. More recent research aligns the querying process with a Markov decision process and learns to select the long-term best query with contextual bandit~\cite{dong2023active} or reinforcement learning~\cite{zha2020meta}. Unlike semi-supervised labeling, which requires one-time human participation in the initial stage, active learning needs a continuous supply of information from humans to adaptively select queries.

\emph{Data programming.}
Data programming~\cite{ratner2016data,ratner2017snorkel} is a weakly-supervised approach that infers labels based on human-designed labeling functions. The labeling functions are often some heuristic rules and vary for different data types, e.g., seed words for text classification~\cite{zha2019multi}, masks for image segmentation~\cite{hooper2021cut}, etc. However, sometimes the labeling functions may not align with human intentions. To address this limitation, researchers have proposed interactive data programming~\cite{boeckinginteractive,galhotra2021adaptive}, where humans participate more by interactively providing feedback to refine labeling functions. Data programming methods often require minimum human participation or, at most, partial participation. Thus, the methods in this research line are often more desirable when we need to quickly generate a large number of labels.

\emph{Distant supervision.}
Another weakly-supervised approach is distant supervision, which assigns labels by leveraging external sources. A famous application of distant supervision is on relation extraction~\cite{mintz2009distant}, where the semantic relationships between entities in the text are labeled based on external data, such as Freebase~\cite{bollacker2008freebase}. Distant supervision is often an automated approach that does not require human participation. However, the automatically generated labels can be noisy if there is a discrepancy between the dataset and the external source.

\subsubsection{Challenges.}
The main challenge for data labeling stems from striking a balance between label quality, label quantity, and financial cost. If given adequate financial support, it is possible to hire a sufficient number of expert annotators to obtain a satisfactory quantity of high-quality labels. However, when we have a relatively tight budget, we often have to resort to more efficient labeling strategies. Identifying the proper labeling strategy often requires domain knowledge to balance different tradeoffs, particularly human labor and label quality/quantity. Another difficulty lies in the subjectivity of labeling. While the instructions may be clear to the designer, they may be misinterpreted by annotators, which leads to labeling noise. Last but not least, ethical considerations, such as data privacy and bias, remain a pressing issue, especially when the labeling task is distributed to a large and undefined group of people.

\subsection{Data Preparation}
\label{sec:3:3}

Data preparation involves cleaning and transforming raw data into a format that is appropriate for model training. Conventionally, this process often necessitates a considerable amount of engineering work with laborious trial and error. To automate this process, state-of-the-art approaches often adopt search algorithms to discover the most effective strategies. In this subsection, we introduce the need, representative methods, and challenges for data preparation.

\subsubsection{Need for Data Preparation} Raw data is often not ready for model training due to potential issues such as noise, inconsistencies, and unnecessary information, leading to inaccurate and biased results. For instance, the model could overfit on noises, outliers, and irrelevant extracted features, resulting in reduced generalizability~\cite{ying2019overview}. If sensitive information (e.g., race and gender) is not removed, the model may unintentionally learn to make biased predictions~\cite{wan2022processing}. In addition, the raw feature values may negatively affect model performance if they are in different scales or follow skewed distributions~\cite{ahsan2021effect}. Thus, it is imperative to clean and transform data. The need can also be verified by a Forbes survey~\cite{press_2022}, which suggests that data preparation accounts for roughly 80\% of the work of data scientists.

\subsubsection{Methods} We will review and discuss the techniques for achieving three key data preparation objectives, namely data cleaning, feature extraction, and feature transformation.

\emph{Data cleaning.} Data cleaning is the process of identifying and correcting errors, inconsistencies, and inaccuracies in datasets. Traditional methods repair data with programmatic automation, e.g., imputing missing values with mean or median~\cite{zhang2016missing} and scanning all data to find duplicates. However, such heuristics can be inaccurate or inefficient. Thus, learning-based methods have been developed, such as training a regression model to predict missing values~\cite{lakshminarayan1996imputation}, efficiently estimating the duplicates with sampling~\cite{heise2014estimating}, and correcting labeling errors~\cite{jiang2022information}.  Contemporary data cleaning methods often do not solely focus on the cleaning itself, but rather on learning to improve final model performance. For instance, a recent study has adopted search algorithms to automatically identify the best cleaning strategy to optimize validation performance~\cite{krishnan2019alphaclean}. Beyond automation, researchers have studied collaboration-oriented cleaning methods. For example, a hybrid human-machine workflow is proposed to identify duplicates by presenting similar pairs to humans for annotation~\cite{wang2012crowder}.

\emph{Feature extraction.} Feature extraction is an important step in extracting relevant features from raw data. For training traditional machine learning models, we often need to extract features based on domain knowledge of the data type being targeted. Common features used for images include color features, texture features, intensity features, etc.~\cite{salau2019feature}. For time series data, temporal, statistical, and spectral features are often considered~\cite{barandas2020tsfel}. Deep learning, in contrast, automatically extracts features by learning the weights of neural networks, which requires less domain knowledge. For instance, convolutional neural networks can be used in both images~\cite{krizhevsky2017imagenet} and time series~\cite{wang2017time}. The boundary between data and model becomes blurred with deep learning feature extractors, which operate on the data while also being an integral part of the model. Although deep extractors could learn high-quality feature representations, the extraction process is uninterpretable and may amplify the bias in the learned representation~\cite {wan2022processing}. Therefore, traditional feature extraction methods are often preferred in high-stakes domains for interpretability and removing sensitive information.

\emph{Feature transformation.} Feature transformation refers to the process of converting the original features into a new set of features, which can often lead to improved model performance. Some typical transformations include normalization, which scales the feature into a bounding range, and standardization, which transforms features so that they have a mean of zero and a standard deviation of one~\cite{ali2014data}. Other strategies include log transformation and polynomial transformation to smooth the long-tail distribution and create new features through multiplication~\cite{bisong2019introduction}. These transformation methods can be combined in different ways to improve model performance. For example, a representative work builds a transformation graph for a given dataset, where each node is a type of transformation, and adopts reinforcement learning to search for the best transformation strategy~\cite{khurana2018feature}. Learning-based methods often yield superior performance by optimizing transformation strategies based on the feedback obtained from the model.

\subsubsection{Challenges.}
Properly cleaning and transforming data is challenging due to the unique characteristics of different datasets. For example, the errors and inconsistencies in text data are quite different from those in time-series data. Even if two datasets have the same data type, their feature values and potential issues can be very diverse. Thus, researchers and data scientists often need to devote a significant amount of time and effort to clean the data. Although learning-based methods can search for the optimal preparation strategy automatically~\cite{khurana2018feature,krishnan2019alphaclean}, it remains a challenge to design the appropriate search space, and the search often requires a non-trivial amount of time.

\subsection{Data Reduction}
\label{sec:3:4}

The goal of data reduction is to reduce the complexity of a given dataset while retaining its essential information. This is often achieved by either reducing the feature size or the sample size. Our discussion will focus on the need for data reduction, representative methods for feature and sample size reduction, and challenges.

\subsubsection{Need for Data Reduction}
With more data being collected at an unprecedented pace, data reduction plays a critical role in boosting training efficiency. From the sample size perspective, reducing the number of samples leads to a simpler yet representative dataset, which can alleviate memory and computation constraints. It also helps to alleviate data imbalance issues by downsampling the samples from the majority class~\cite{prusa2015using}. Similarly, reducing feature size brings many benefits. For example, eliminating irrelevant or redundant features mitigates the risk of overfitting~\cite{li2017feature}. Smaller feature sizes will also enable faster training and inference in model deployment~\cite{wang2022bed}. In addition, only keeping a subset of features will make the model more interpretable~\cite{chuang2023efficient,wang2022accelerating,chuangcortx}. Data reduction techniques can enable the model to focus only on the essential information, thereby enhancing accuracy, efficiency, and interpretability.

\subsubsection{Methods for Reducing Feature Size.} From the feature perspective, we discuss two common reduction strategies.

\emph{Feature selection.} Feature selection is the process of selecting a subset of features most relevant to the intended tasks~\cite{li2017feature}. It can be broadly classified into filter, wrapper, and embedded methods. Filter methods~\cite{thaseen2017intrusion} evaluate and select features independently using a scoring function based on statistical properties such as information gain~\cite{azhagusundari2013feature}. Although filter methods are very efficient, they ignore feature dependencies and interactions with the model. Wrapper methods alleviate these issues by leveraging the model performance to assess the quality of selected features and refining the selection iteratively~\cite{yan2015feature}. While these methods often achieve better performances, they are computationally more expensive. Embedded methods, from another angle, integrate feature selection into the model training process~\cite{wang2015embedded} so that the selection process is optimized in an end-to-end manner. Beyond automation, active feature selection takes into account human knowledge and incrementally selects the most appropriate features~\cite{zhang2019active,schnapp2021active}. Feature selection reduces the complexity, producing cleaner and more understandable data while retaining feature semantics.

\emph{Dimensionality reduction.} Dimensionality reduction aims to transform high-dimensional features into a lower-dimensional space while preserving the most representative information. The existing methods can be mainly categorized into linear and non-linear techniques. The former generates new features via linear combinations of features from the original data. One of the most popular algorithms is Principal Component Analysis (PCA)~\cite{abdi2010principal}, which performs orthogonal linear combinations of the original features based on the variance in an unsupervised manner. Another representative method targeted for supervised scenarios is Linear Discriminant Analysis (LDA)~\cite{xanthopoulos2013linear}, which statistically learns linear feature combinations that can separate classes well. Linear techniques, however, may not always perform well, especially when features have complex and non-linear relationships. Non-linear techniques address this issue by utilizing nonlinear mapping functions. A popular technique is autoencoders~\cite{bank2020autoencoders}, which use neural networks to encode the original features into a low-dimensional space and reconstruct the features using a neural decoder.

\subsubsection{Methods for Reducing Sample Size} The reduction of samples is typically achieved with \emph{instance selection}, which selects a representative subset of data samples that retain the original properties of the dataset. The existing studies can be divided into wrapper and filter methods. The former selects instances based on scoring functions. For example, a common strategy is to select border instances since they can often shape the decision boundary~\cite{riquelme2003finding}. Wrapper methods, in contrast, select instances based on model performance~\cite{sutton2012introduction}, which considers the interaction effect with the model. Instance selection techniques can also alleviate data imbalance issues by undersampling the majority class, e.g., with random undersampling~\cite{prusa2015using}. More recent work adopts reinforcement learning to learn the best undersampling strategies~\cite{liu2020mesa}. Overall, instance selection is a simple yet effective way to reduce data sizes or balance data distributions.

\subsubsection{Challenges}
The challenges of data reduction are two-folded. On the one hand, selecting the most representative data or projecting data in a low-dimensional space with minimal information loss is non-trivial. While learning-based methods can partially address these challenges, they may necessitate substantial computational resources, especially when dealing with extremely large datasets, e.g., the wrapper and reinforcement learning methods~\cite{yan2015feature,sutton2012introduction,liu2020mesa}. Therefore, achieving both high accuracy and efficiency is challenging. On the other hand, data reduction can potentially amply data bias, raising fairness concerns. For example, the selected features could be over associating with protected attributes~\cite{xing2021fairness}. Fairness-aware data reduction is a critical yet under-explored research direction.

\subsection{Data Augmentation} 
\label{sec:3:5}

Data augmentation is a technique to increase the size and diversity of data by artificially creating variations of the existing data, which can often improve the model performance. It is worth noting that even though data augmentation and data reduction seem to have contradictory objectives, they can be used in conjunction with each other. While data reduction focuses on eliminating redundant information, data augmentation aims to enhance data diversity.  We will delve into the need for data augmentation, various representative methods, and the associated challenges.

\subsubsection{Need for Data Augmentation}
Modern machine learning algorithms, particularly deep learning, often require large amounts of data to learn effectively. However, collecting large datasets, especially annotated data, is labor-intensive. By generating similar data points with variance, data augmentation helps to expose the model to more training examples, hereby improving accuracy, generalization capabilities, and robustness. Data augmentation is particularly important in applications where there is limited data available. For example, it is often expensive and time-consuming to acquire well-annotated medical data~\cite{chlap2021review}. Data augmentation can also alleviate class imbalance issues, where there is a disproportionate ratio of training samples in each class, by augmenting the data from the under-represented class.

\subsubsection{Common Augmentation Methods} In general, data augmentation methods often manipulate the existing data to generate variances or synthesize new data. We discuss some representative methods in each category below.

\emph{Basic manipulation.} This research line involves making minor modifications to the original data samples to produce augmented samples directly. Various strategies have been proposed in the computer vision domain, such as scaling, rotation, flipping, and blurring~\cite{zhang2015character}. One notable approach is Mixup~\cite{zhang2018mixup}, which interpolates the existing data samples to create new samples. It is shown that Mixup serves as a regularizer, encouraging the model to prioritize simpler linear patterns, which in turn enhances the generation performance~\cite{zhang2018mixup}. More recent studies use learning-based algorithms to automatically search for augmentation strategies. A representative work is AutoAugment, which uses reinforcement learning to iteratively improve the augmentation policies~\cite{cubuk2019autoaugment}. Beyond image data, basic manipulation often needs to be tailored for the other data types, such as permutation and jittering in time-series data~\cite{wen2021time}, mixing data in the hidden space for text data to retain semantic meanings~\cite{chen2020mixtext}, and mixing graphon for graph data~\cite{han2022g}.

\emph{Augmentation data synthesis.} Another category focuses on synthesizing new training samples by learning the distribution of the existing data, which is typically achieved by generative modeling. GAN~\cite{goodfellow2020generative,zhang2019self} has been widely used for data augmentation~\cite{frid2018synthetic}. The key idea is to train a discriminator in conjunction with a generator, making the latter generate synthetic data that closely resembles the existing data. GAN-based data augmentation has also been used to augment other data types, such as time-series data~\cite{li2022tts} and text data~\cite{shorten2021text}. Other studies have used Variational Autoencoder~\cite{hsu2017unsupervised} and diffusion models~\cite{ho2022cascaded} to achieve augmentation. Compared to basic manipulation that augments data locally, data synthesis learns data patterns from the global view and generates new samples with a learned model.

\subsubsection{Methods Tailored for Class Imbalance} Class imbalance is a fundamental challenge in machine learning, where the number of majority samples is much larger than that of minority samples. Data augmentation can be used to perform \emph{upsampling} on the minority class to balance the data distribution. One popular approach is SMOTE~\cite{chawla2002smote}, which involves generating synthetic samples by linearly interpolating between minority instances and their neighbors. ADASYN~\cite{he2008adasyn} is an extension of SMOTE that generates additional synthetic samples for data points that are more difficult to learn, as determined by the ratio of majority class samples in their nearest neighbors. A recent study proposes AutoSMOTE, a learning-based algorithm that searches for best oversampling strategies with reinforcement learning~\cite{zha2022towards}.

\subsubsection{Challenges}
One critical challenge in data augmentation is that there is no single augmentation strategy that is suitable for all scenarios. Different data types may require diverse strategies. For example, compared to image data, graph data is irregular and not well-aligned, and thus the vanilla Mixup strategy can not be directly applied~\cite{han2022g}. Even though two datasets have the same data type, the optimal strategy differs. For instance, we often need to upsample the minority samples differently to achieve the best results~\cite{zha2022towards}. Although search-based algorithms can identify the best strategies with trial and error, it also increases the computation and storage costs, which can be a limiting factor in some applications. More effective and efficient data augmentation techniques are required to overcome these challenges.

\subsection{Pipeline Search}
\label{sec:3:6}
In real-world applications, we often encounter complex data pipelines, where each pipeline step corresponds to a task associated with one of the aforementioned sub-goals. Despite the progress made in each individual task, a pipeline typically functions as a whole, and the various pipeline steps may have an interactive effect. For instance, the best data augmentation strategy may depend on the selected features. Pipeline search is a recent trend that tries to automatically search for the best combinations. This subsection introduces some representative pipeline search algorithms.

One of the first pipeline search frameworks is AutoSklearn~\cite{feurer2015efficient}. It performs a combined search of preprocessing modules, models, and the associated hyperparameters to optimize the validation performance. However, they use a very small search space for preprocessing modules. DARPA’s Data-Driven Discovery of Models (D3M) program pushes the progress further by building an infrastructure for pipeline search~\cite{milutinovic2020evaluation}. Although D3M originally focused on automated model discovery, it has developed numerous data-centric modules for processing data. Building upon D3M, AlphaD3M uses Monte-Carlo Tree Search to identify the best pipeline~\cite{drori2021alphad3m}. D3M is then tailored for time-series anomaly detection~\cite{lai2021tods} and video analysis~\cite{zha2021autovideo}. Deepline enables the search within a large number of data-centric modules using multi-step reinforcement learning~\cite{heffetz2020deepline}. ClusterP3S allows for personalized pipelines to be created for various features, utilizing clustering techniques to enhance search efficiency~\cite{martinez2023towards}.

Despite these progresses, pipeline search still faces a significant challenge due to the high computational overhead since the search algorithm often needs to try different module combinations repeatedly. This overhead becomes more pronounced as the number of modules increases, leading to an exponential growth of the search space. Thus, more efficient search strategies~\cite{heffetz2020deepline,martinez2023towards} are required to enable a broader application of pipeline search in real-world scenarios.

%% file: sec4.tex
\begin{figure}[t]
  \centering
    \includegraphics[width=1.0\textwidth]{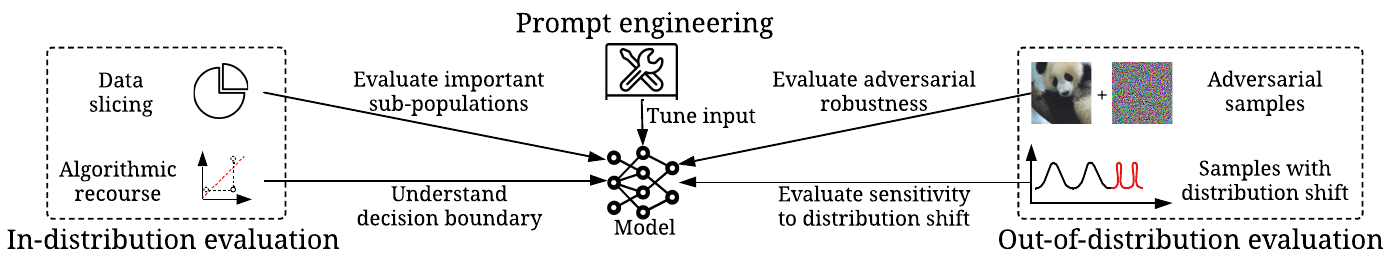}
  \caption{An overview of inference data development.}
  \label{fig:inferencedevelopmentoverview}
\end{figure}

\section{Inference Data Development}
\label{sec:4}
Another crucial component in building AI systems is to design inference data to evaluate a trained model or unlock a specific capability of the model. In the conventional model-centric paradigm, we often adopt a hold-out evaluation set that is not included in the training data to measure model performance using specific metrics such as accuracy. However, relying solely on performance metrics may not fully capture many important properties of a model, such as its robustness, generalizability, and rationale in decision-making. Moreover, as models become increasingly large, it becomes possible to obtain the desired predictions by solely engineering the data input. This section introduces some representative methods that evaluate models from a more granular view, or engineering data inputs for inference, shown in Figure~\ref{fig:inferencedevelopmentoverview}. Our discussion involves in-distribution set evaluation (Section~\ref{sec:41}), out-of-distribution evaluation (Section~\ref{sec:42}), and prompt engineering (Section~\ref{sec:43}).
We summarize the relevant tasks and methods in Table~\ref{tbl:inferencemethodsummary}.


\subsection{In-distribution Evaluation} 
\label{sec:41}

In-distribution evaluation data construction aims to generate samples that conform to training data. We will begin by addressing the need for constructing in-distribution evaluation sets. Next, we will review representative methods for two scenarios: evaluating important sub-populations on which the model underperforms through data slicing, and assessing decision boundaries through algorithmic recourse. Lastly, we will discuss the challenges.



\subsubsection{Need for In-distribution Evaluation}

In-distribution evaluation is the most direct way to assess the quality of trained models, as it reflects their capabilities within the training distribution. The need for a more fine-grained in-distribution evaluation is two-fold. Firstly, models that perform well on average may fail to perform adequately on specific sub-populations, requiring identification and calibration of underrepresented groups to avoid biases and errors, particularly in high-stakes applications~\cite{otles2021mind,meng2022interpretability}.
Secondly, it is crucial to understand the decision boundary and inspect the model ethics before deployment, especially in risky applications like policy making~\cite{souza2019data}.

\subsubsection{Data Slicing}

Data slicing involves partitioning a dataset into relevant sub-populations and evaluating a model's performance on each sub-population separately. A common approach to data slicing is to use pre-defined criteria, such as age, gender, or race~\cite{barenstein2019propublica}. However, data in many real-world applications can be complex, and properly designing the partitioning criteria heavily relies on domain knowledge, such as slicing 3-D seismic data in geophysics~\cite{zeng1998stratal} and program slicing~\cite{santelices2013quantitative}.

To reduce human effort, automated slicing methods have been developed to discover important data slices by sifting through all potential slices in the data space. One representative work is SliceFinder~\cite{chung2019slice}, which identifies slices that are both interpretable (i.e., slicing based on a small set of features) and problematic (the model performs poorly on the slice). To solve this search problem, SliceFinder offers two distinct methods, namely the tree-based search and the lattice-based search. The former is more efficient, while the latter has better efficacy. SliceLine~\cite{sagadeeva2021sliceline} is another notable work that addresses the scalability limitations of slice finding by focusing on both algorithmic and system perspectives. This approach is motivated by frequent itemset mining and leverages relevant monotonicity properties and upper bounds for effective pruning. Moreover, to address hidden stratification, which occurs when each labeled class contains multiple semantically distinct subclasses, GEORGE~\cite{sohoni2020no} employs clustering algorithms to slide data across different subclasses. Another tool for automated slicing is Multiaccuracy~\cite{kim2019multiaccuracy}, where a simple ``auditor'' is trained to predict the residual of the full model using input features. Multiaccuracy, in general, is an efficient approach since it only requires a small amount of audit data. 
Data slicing allows researchers and practitioners to identify biases and errors in a model's predictions and calibrate the model to improve its overall capabilities.

\begin{table}[t]
\centering
\caption{Papers for achieving different sub-goals of inference data development.}
\footnotesize
\setlength{\tabcolsep}{1.0pt}
\label{tbl:inferencemethodsummary}
\begin{tabular}{l|l|l|l|l} \toprule
\multirow{2}{*}{\textbf{Sub-goal}} & \multirow{2}{*}{\textbf{Task}} & \multirow{2}{*}{\textbf{Method type}} &  \multirow{2}{*}{\textbf{\shortstack[l]{Automation level/\\ participation degree}}} & \multirow{2}{*}{\textbf{References}} \\
~ & ~ & ~ & ~ & ~ \\
\midrule

\multirow{4}{*}{\shortstack[l]{In-\\ distribution}} & Data slicing & Collaboration  &  Minimum & \cite{barenstein2019propublica} \\
~ & Data slicing & Collaboration  &  Partial & \cite{zeng1998stratal,santelices2013quantitative} \\
~ & Data slicing & Automation & Learning-based & \cite{chung2019slice,sagadeeva2021sliceline,sohoni2020no,kim2019multiaccuracy} \\
~ & Algorithmic recourse & Collaboration & Minimum &  \cite{kanamori2020dace,carreira2021counterfactual,lucic2022focus,wachter2017counterfactual,dandl2020multi,dhurandhar2019model,sharma2019certifai,laugel2018comparison,poyiadzi2020face,becker2021step,blanchart2021exact} \\
\cline{1-5}

\multirow{6}{*}{\shortstack[l]{Out-of-\\ distribution}} & Adversarial samples & Collaboration &  Minimum & \cite{hendrycks2019benchmarking} \\
~ & Adversarial samples & Automation &  Learning-based & \cite{biggio2013evasion,moosavi2016deepfool,madry2017towards,eykholt2018robust,papernot2017practical,chen2017zoo,shafahi2018poison} \\
~ & Distribution shift & Collaboration &  Full & \cite{ding2021retiring,koh2021wilds,saenko2010adapting} \\
~ & Distribution shift & Collaboration &  Partial & \cite{gu2019using,shankar2021image} \\
~ & Distribution shift & Automation &  Programmatic & \cite{gretton2009covariate,sugiyama2007covariate,lipton2018detecting,azizzadenesheli2019regularized} \\
~ & Distribution shift & Automation &  Learning-based & \cite{farahani2021brief,guan2021domain} \\

\cline{1-5}
\multirow{3}{*}{\shortstack[l]{Prompt\\ engineering}} & Manual engineering & Collaboration & Partial & \cite{schick2020few,schick2020exploiting,schick2020s} \\
~ & Automated engineering & Automation & Programmatic & \cite{jiang2020can,yuan2021bartscore,haviv2021bertese} \\
~ & Automated engineering & Automation & Learning-based & \cite{wallace2019universal,gao2021making} \\

\bottomrule
\end{tabular}
\end{table}

\subsubsection{Algorithmic Recourse} 

Algorithmic recourse (also known as  ``counterfactuals''~\cite{wachter2017counterfactual} in the explainable AI domain) aims to generate a hypothetical set of samples that can flip model decisions toward preferred outcomes. For example, if an individual is denied a loan, algorithmic recourse seeks the closest sample (e.g., with a higher account balance) that would have been approved. Hypothetical samples derived through algorithmic recourse are valuable in understanding decision boundaries. For the previously mentioned example, the hypothetical sample addresses the question of how the individual could have been approved and also aids in the detection of potential biases across individuals.

The existing
methods primarily vary in their strategies for identifying hypothetical samples, and can generally be classified into white-box and black-box methods. White-box methods necessitate access to the evaluated models, which can be achieved through complete internals~\cite{kanamori2020dace,carreira2021counterfactual,lucic2022focus}, gradients~\cite{wachter2017counterfactual}, or solely the prediction function~\cite{dandl2020multi,dhurandhar2019model,sharma2019certifai,laugel2018comparison}. Conversely, black-box methods do not require access to the model at all. For example, Dijkstra’s algorithm is employed to obtain the shortest path between existing training data points to find recourse under certain distributions~\cite{poyiadzi2020face}. An alternative approach involves dividing the feature space into pure regions, where all data points belong to a single class, and utilizing graph traversing techniques~\cite{becker2021step,blanchart2021exact} to identify the nearest recourse. Given that the target label for reasoning is usually inputted by humans, these recourse methods all require minimal human participation.


\subsubsection{Challenges}

The main challenge of constructing in-distribution evaluation sets lies in identifying the targeted samples effectively and efficiently. In the case of data slicing, determining the optimal subset of data is particularly challenging due to the exponential increase in the number of possible subsets with additional data points. Similarly, identifying the closest recourse when limited information is available also requires significant effort.


\subsection{Out-of-distribution Evaluation}
\label{sec:42}
Out-of-distribution evaluation data refers to a set of samples that follow a distribution that differs from the one observed in the training data. We begin by discussing the need for out-of-distribution evaluation, followed by a review of two representative tasks: generating adversarial samples and generating samples with distribution shifts. Then we delve into the challenges associated with out-of-distribution data generation.

\subsubsection{Need for Out-of-distribution Evaluation}
Although modern machine learning techniques generally perform well on in-distribution datasets, the distribution of data in the deployment environment may not align with the training data~\cite{shen2021towards}. Out-of-distribution evaluation primarily assesses a model's ability to generalize to unexpected scenarios by utilizing data samples that differ significantly from the ones used during training. This evaluation can uncover the transferability of a model and instill confidence in its performance in unexpected scenarios. Out-of-distribution evaluation can also provide essential insights into a model's robustness, exposing potential flaws that must be addressed before deployment. This is crucial in determining whether the model is secure in real-world deployments.

\subsubsection{Generating Adversarial Samples}

Adversarial samples are the ones with intentionally manipulated or modified input data in a way that causes a model to make incorrect predictions. Adversarial samples can aid in comprehending a model's robustness and are typically generated by applying perturbations to the input data. Manual perturbation involves adding synthetic and controllable perturbations, such as noise and blur, to the original data~\cite{hendrycks2019benchmarking}.

Automated methods design learning-based strategies to generate perturbations automatically and are commonly classified into four categories: white-box attacks, physical world attacks, black-box attacks, and poisoning attacks. White-box attacks involve the attacker being provided with the model and victim sample. Examples of white-box attacks include Biggio's attack~\cite{biggio2013evasion}, DeepFool~\cite{moosavi2016deepfool}, and projected
gradient descent attack~\cite{madry2017towards}. Physical world attacks involve introducing real perturbations to real-world objects. For instance, in the work by~\cite{eykholt2018robust}, stickers were attached to road signs to significantly impact the sign identifiers of autonomous cars. Black-box attacks are often applied when an attacker lacks access to a classifier's parameters or training set but possesses information regarding the data domain and model architecture. In~\cite{papernot2017practical}, the authors exploit the transferability property to generate adversarial examples. A zero-th order optimization-based black-box attack is proposed in \cite{chen2017zoo} that leverages the prediction confidence for the victim sample. Poisoning attacks involve the creation of adversarial examples prior to training, utilizing knowledge about model architectures. For instance, the poison frogs technique~\cite{shafahi2018poison} inserts an adversarial image into the training set with a true label. By evaluating a trained model on various adversarial samples, we can gain a better understanding of the potential weaknesses of the model in deployment. This can help us take steps to prevent undesirable outcomes.

\subsubsection{Generating Samples with Distribution Shift} 

Generating samples with distribution shifts enables the evaluation of a model on a different distribution. One straightforward way is to collect data with varying patterns, such as shifts across different times or locations~\cite{ding2021retiring}, camera traps for wildlife monitoring~\cite{koh2021wilds}, and diverse domains~\cite{saenko2010adapting}. A more efficient approach would involve constructing the evaluation set from pre-collected data. To illustrate, some studies~\cite{gu2019using,shankar2021image} generate various sets of contiguous video frames that appear visually similar to humans but lead to inconsistent predictions due to the small perturbations.

Apart from natural distribution shifts in real-world data, synthetic distribution shifts are widely adopted, including three types: 1) covariate shift, which assumes that the input distribution is shifted~\cite{gretton2009covariate,sugiyama2007covariate}, 2) label shift, which assumes that the label distribution is shifted~\cite{lipton2018detecting,azizzadenesheli2019regularized}, and 3) general distribution shift, which assumes that both the input and label distributions are shifted~\cite{farahani2021brief,guan2021domain}. Biased data sampling can be used to synthesize covariate shifts or label shifts, whereas learning-based methods are typically required to synthesize general distribution shifts~\cite{farahani2021brief,guan2021domain}. Generating samples with distribution shift is essential in evaluating a model's transferability, especially when there is a distribution gap between the training and deployment environments.

\subsubsection{Challenges}

The challenges for out-of-distribution generation set construction are two-fold. Firstly, generating high-quality out-of-distribution data is challenging. If the training data is not representative, it may be difficult to generate appropriate data. Furthermore, the generation models may encounter mode collapse issues, meaning that they only generate a limited number of similar samples and disregard the diversity of the target distribution. Secondly, evaluating the quality of out-of-distribution generation is difficult since no single metric can capture the diversity and quality of the generated samples. Commonly used metrics, such as likelihood or accuracy, may not be suitable as they may exhibit bias toward generating samples similar to the training data. Therefore, various evaluation metrics have been proposed to assess the distance between in-distribution and out-of-distribution samples~\cite{sangkloy2017scribbler,borgwardt2006integrating,obukhov2020quality,betzalel2022study,jiang2023weight}. Overall, creating high-quality out-of-distribution data is a complex and demanding task that requires meticulous design.

\subsection{Prompt Engineering}
\label{sec:43}

With the advent of large language models, it becomes feasible to accomplish a task by solely fine-tuning the input to probe knowledge from the model, while keeping the model fixed. Prompt engineering is an emerging task that aims to design and construct high-quality prompts to achieve the most effective performance on downstream tasks~\cite{liu2023pre}. For example, when performing text summarization, we can provide the texts we want to summarize followed by specific instructions such as "summarize it" or "TL;DR" to guide the inference. Prompt engineering revolutionizes the traditional workflow by fine-tuning the input data rather than the model itself to achieve a given task.

A natural way is to perform \emph{manual prompt engineering} by creating templates. For example, in~\cite{schick2020few,schick2020exploiting,schick2020s}, the authors have pre-defined templates for few-shot learning in text classification and conditional text generation tasks. However, manually crafting templates may not be sufficient to discover the optimal prompts for complex tasks. Thus, \emph{automated prompt engineering} has been studied. Common programmatic approaches include mining the templates from an external corpus~\cite{jiang2020can} and paraphrasing with a seed prompt~\cite{yuan2021bartscore,haviv2021bertese}. Learning-based methods automatically generate the prompt tokens by gradient-based search~\cite{wallace2019universal} or generative models~\cite{gao2021making}. The primary obstacle in prompt engineering arises from the absence of a universal prompt template that consistently performs well. Various templates may result in different model behaviors, and obtaining the desired answers is not guaranteed. Therefore, further research is necessary to gain insight into the response of the model to prompts and guide the prompt engineering process.

%% file: sec5.tex
\section{Data Maintenance}
\label{sec:5}
\begin{figure}[t]
  \centering
    \includegraphics[width=1.0\textwidth]{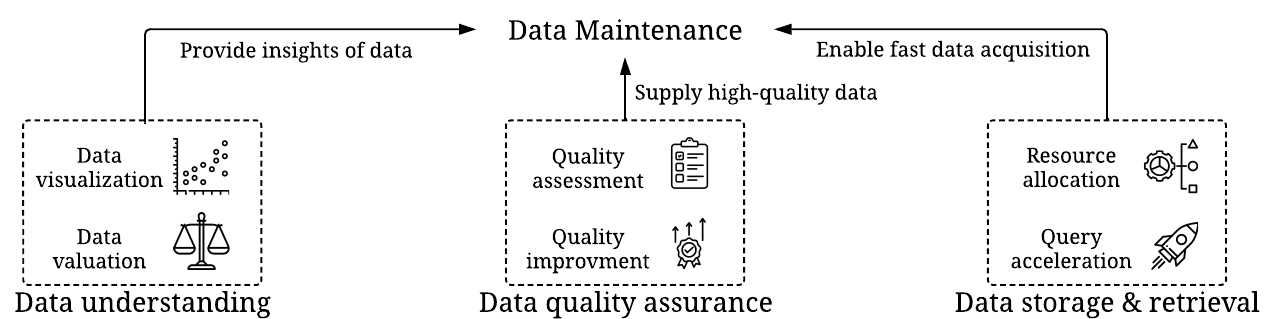}
  \caption{An overview of data maintenance.}
  \label{fig:datamaintenanceoverview}
\end{figure}

In production scenarios, data is not created once but is rather continuously updated, making data maintenance a significant challenge that must be considered to ensure reliable and instant data supply in building AI systems. This section provides an overview of the need, representative methods (as depicted in Figure~\ref{fig:datamaintenanceoverview}), and challenges of data maintenance. Our discussion spans across three aspects: data understanding (Section~\ref{sec:51}), data quality assurance (Section~\ref{sec:52}), and data storage \& retrieval (Section~\ref{sec:53}). Additionally, Table~\ref{tbl:maintainmethodsummary} summarizes the relevant tasks and methods.

\subsection{Data Understanding}
\label{sec:51}

To ensure proper maintenance, it is essential to first understand the data. The following discussion covers the need for data understanding techniques, ways to gain insights through visualization and valuation, and the challenges involved.

\subsubsection{Need for Data Understanding Techniques.}
Real-world data often comes in large volumes and complexity, which can make it difficult to understand and analyze. There are three main reasons why data understanding techniques are crucial. Firstly, comprehending a large number of raw data samples can be challenging for humans. To make it more manageable, we need to summarize the data and present it in a more concise and accessible way. Secondly, real-world data is often high-dimensional, while human perception is limited to two-or-three-dimensional space. Therefore, visualizing data in a lower-dimensional space is essential for understanding the data. Finally, it is crucial for organizations and stakeholders to understand the value of their data assets and the contribution of each data sample to the performance.

\subsubsection{Data Visualization}

Human beings are visual animals, and as such, we have a natural tendency to process and retain information presented in a pictorial and graphical format. Data visualization aims to leverage this innate human trait to help us better understand complex data.
In what follows, we will discuss three relevant research topics: visual summarization, clustering for visualization, and visualization recommendation.


\emph{Visual summarization.} Summarizing the raw data as a set of graphical diagrams can assist humans in gaining insights through a condensed interface. Despite its wide application, generating a faithful yet user-friendly summarization diagram is a non-trivial task. For example, it is hard to select the right visualization format. Radial charts (e.g., star glyphs and rose charts) and linear charts (e.g., line charts and bar charts) are two common formats for visualization. However, it is controversial which format is better. Although empirical evidence suggests that linear charts are superior to radial charts for many analytical tasks~\cite{burch2014benefits}, radial charts are often more natural and memorable~\cite{borkin2013makes}. In some cases, it is acceptable to compromise on the faithfulness of data representation in favor of enhanced memorability or space efficiency~\cite{burch2014benefits,waldner2019comparison}. For readers who are interested, \cite{desnoyers2011toward} and \cite{franconeri2021science} provide a comprehensive taxonomy of visualization formats. Although automated scripts can generate plots, the process of visual summarization often demands minimal human participation to select the most appropriate visualization formats.

\begin{table}[t]
\centering
\caption{Papers for achieving different sub-goals of data maintenance.}
\label{tbl:maintainmethodsummary}
\footnotesize
\begin{tabular}{l|l|l|l|l} \toprule
\multirow{2}{*}{\textbf{Sub-goal}} & \multirow{2}{*}{\textbf{Task}} & \multirow{2}{*}{\textbf{Method type}} &  \multirow{2}{*}{\textbf{\shortstack[l]{Automation level/\\ participation degree }}} & \multirow{2}{*}{\textbf{Reference}} \\
~ & ~ & ~ & ~ & ~ \\
\midrule

\multirow{6}{*}{Understanding} & Visual summarization & Collaboration & Minimum & \cite{burch2014benefits,borkin2013makes,waldner2019comparison,desnoyers2011toward,franconeri2021science} \\
~ & Clustering for visualization & Automation & Learning-based & \cite{fahad2014survey} \\
~ & Visualization recommendation & Automation & Programmatic & \cite{wongsuphasawat2015voyager} \\
~ & Visualization recommendation & Automation & Learning-based & \cite{luo2018deepeye} \\
~ & Visualization recommendation & Collaboration & Partial & \cite{shen2021towardsnli,srinivasan2021snowy} \\
~ & Valuation & Automation & Learning-based & \cite{ghorbani2020distributional,agarwal2019marketplace,ghorbani2019data} \\
\cline{1-5}

\multirow{5}{*}{Quality assurance} & Quality assessment & Collaboration & Minimum/partial & \cite{sadiq2018data,xue2022knowledge,batini2009methodologies,pipino2002data} \\
~ & Quality improvement & Automation & Programmatic &  \cite{basu1995discovering,chu2013discovering,bohannon2006conditional}\\
~ & Quality improvement & Automation & Learning-based & \cite{baylor2017tfx} \\
~ & Quality improvement & Automation & Pipeline & \cite{schelter2018automating,thirumuruganathan2020data} \\
~ & Quality improvement & Collaboration & Partial & \cite{gamboa2022human,deodhar2022human,wang2021crowdsourcing,chen2020building} \\
\cline{1-5}

\multirow{6}{*}{Storage \& retrieval} & Resource allocation & Automation &  Programmatic & \cite{apache,yarn,white2012hadoop} \\
~ & Resource allocation & Automation & Learning-based & \cite{herodotou2011starfish,van2017automatic}\\
~ & Query index selection & Automation & Programmatic & \cite{sun2019end,chaudhuri1997efficient,valentin2000db2} \\
~ & Query index selection & Automation & Learning-based & \cite{pedrozo2018adaptive,sadri2020online} \\
~ & Query rewriting & Automation & Programmatic & \cite{baik2019bridging,chavan2011dbridge} \\
~ & Query rewriting & Automation & Learning-based & \cite{he2016learning,zhou2021dbmind} \\

\bottomrule
\end{tabular}
\end{table}

\emph{Clustering for visualization.} Real-world data can be high-dimensional and with complex manifold structures. As such, dimensionality reduction techniques (mentioned in Section~\ref{sec:3:4}) are often applied to visualize data in a two-or-three-dimensional space. Furthermore, automated clustering methods~\cite{fahad2014survey} are frequently combined with dimensionality reduction techniques to organize data points in a grouped, categorized, and often color-coded fashion, facilitating human comprehension and insightful analysis of the data.

\emph{Visualization recommendation.} Building upon various visualization formats, there has been a surge of interest  in visualization recommendation, which involves suggesting the most suitable visualization formats for a particular user. Programmatic automation approaches rank visualization candidates based on predefined rules composed of human perceptual metrics such as data type, statistical information, and human visual preference~\cite{wongsuphasawat2015voyager}. Learning-based approaches exploit various machine learning techniques to rank the visualization candidates. An example of such a method is DeepEye~\cite{luo2018deepeye}, which utilizes the statistical information of the data as input and optimizes the normalized discounted cumulative gain (NDCG) based on the quality of the match between the data and the chart. Collaborative visualization techniques allow for a more adaptable user experience by enabling users to continuously provide feedback and requirements for the visualization~\cite{shen2021towardsnli}. A recent study, Snowy~\cite{srinivasan2021snowy} accepts human language as input and generates recommendations for utterances during conversational visual analysis. As visualizations are intended for human users, allowing for human-in-the-loop feedback is crucial in developing visualization recommender systems.

\subsubsection{Data Valuation}

The objective of data valuation is to understand how each data point contributes to the final performance. Such information not only provides valuable insights to stakeholders but is also useful in buying or selling data points in the data market and credit attribution~\cite{ghorbani2020distributional}. To accomplish this, researchers estimate the Shapley value of the data points, which assigns weights to each data point based on its contribution~\cite{agarwal2019marketplace,ghorbani2019data}. A subsequent study has enhanced the robustness of this estimation across multiple datasets and models~\cite{ghorbani2020distributional}. Since calculating the exact Shapley value can be computationally expensive, especially when dealing with a large number of data points, the above methods all adopt learning-based algorithms for efficient estimation.



\subsubsection{Challenges} 
There are two major challenges. Firstly, the most effective data visualization formats and algorithms (e.g., clustering algorithms) are often specific to the domain and influenced by human behavior, making it difficult to select the best option. This selection process often requires human input. Determining how to best interact with humans adds an additional complexity. Secondly, developing efficient data valuation algorithms is challenging, since estimating the Shapley value can be computationally expensive, especially as data sizes continue to grow. Additionally, the Shapley value may only offer a limited perspective on data value, as there are many other important factors beyond model performance, such as the problems that can be addressed through training a model on the data.

\subsection{Data Quality Assurance}
\label{sec:52}
To ensure a reliable data supply, it is essential to maintain data quality. We will discuss why quality assurance is necessary, the key tasks involved in maintaining data quality (quality assessment and improvement), and the challenges.

\subsubsection{Need for Data Quality Assurance}

In real-world scenarios, data and the corresponding infrastructure for data processing are subject to frequent and continuous updates. As a result, it is important not only to create high-quality training or inference data once but also to maintain their excellence in a dynamic environment. Ensuring data quality in such a dynamic environment involves two aspects. Firstly, continuous monitoring of data quality is necessary. Real-world data in practical applications can be complex, and it may contain various anomalous data points that do not align with our intended outcomes. As a result, it is crucial to establish quantitative measurements that can evaluate data quality. Secondly, if a model is affected by low-quality data, it is important to implement quality improvement strategies to enhance data quality, which will also lead to improved model performance.

\subsubsection{Quality Assessment}

Quality assessment develops evaluation metrics to measure the quality of data and detect potential flaws and risks. These metrics can be broadly categorized as either objective or subjective assessments~\cite{sadiq2018data,xue2022knowledge,batini2009methodologies,pipino2002data}. Although objective and subjective assessments may require different degrees of human participation, both of them are used in each paper we surveyed. Thus, we tag each paper with more than one degree of human participation in Table~\ref{tbl:maintainmethodsummary}. We will discuss these two types of assessments in general and provide some representative examples of each.

Objective assessments directly measure data quality using inherent data attributes that are independent of specific applications. Examples of such metrics include accuracy, timeliness, consistency, and completeness. Accuracy refers to the correctness of obtained data, i.e., whether the obtained data values align with those stored in the database. Timeliness assesses whether the data is up-to-date. Consistency refers to the violation of semantic rules defined over a set of data items. Completeness measures the percentage of values that are not null. All of these metrics can be collected directly from the data, requiring only minimal human participation to specify the calculation formula.

Subjective assessments evaluate data quality from a human perspective, often specific to the application and requiring external analysis from experts. Metrics like trustworthiness, understandability, and accessibility are often assessed through user studies and questionnaires. Trustworthiness measures the accuracy of information provided by the data source. Understandability measures the ease with which users can comprehend collected data, while accessibility measures users' ability to access the data. Although subjective assessments may not directly benefit model training, they can facilitate easier collaboration within an organization and provide long-term benefits. Collecting these metrics typically requires full human participation since they are often based on questionnaires.

\subsubsection{Quality Improvement}
Quality improvement involves developing strategies to enhance the quality of data at various stages of a data pipeline. Initially, programmatic automation methods are used to enforce quality constraints, including integrity constraints~\cite{basu1995discovering}, denial constraints~\cite{chu2013discovering}, and conditional functional dependencies~\cite{bohannon2006conditional} between columns. More recently, machine learning-based automation approaches have been developed to improve data quality. For instance, in~\cite{baylor2017tfx}, a data validation module trains a machine learning model on a training set with expected data schema and generalizes it to identify potential problems in unseen scenarios. Furthermore, pipeline automation approaches have been developed to systematically curate data in multiple stages of the data pipeline, such as data integration and data cleaning~\cite{schelter2018automating,thirumuruganathan2020data}. 

Apart from automation, collaborative approaches have been developed to encourage expert participation in data improvement. For example, in autonomous driving~\cite{gamboa2022human} and video content reviewing~\cite{deodhar2022human}, human annotations are continuously used to enhance the quality of training data with the assistance of machine learning models. Moreover, UniProt~\cite{wang2021crowdsourcing}, a public database for protein sequence and function literature, has created a systematic submission system to harness collective intelligence~\cite{chen2020building} for data improvement. This system automatically verifies meta-information, updated versions, and research interests of the submitted literature. All of these methods necessitate partial human participation, as humans must continuously provide information through annotations or submissions.

\subsubsection{Challenges}

Ensuring data quality poses two main challenges. Firstly, selecting the most suitable assessment metric is not a straightforward task and heavily relies on domain knowledge. A single metric may not always be adequate in a constantly evolving environment. Secondly, quality improvement is a vital yet laborious process that necessitates careful consideration. Although automation is crucial in ensuring sustainable data quality, human involvement may also be necessary to ensure that the data quality meets human expectations. Therefore, data assessment metrics and data improvement strategies must be thoughtfully designed.

\subsection{Data Storage \& Retrieval}
\label{sec:53}

Data storage and retrieval systems play an indispensable role in providing the necessary data to build AI systems. To expedite the process of data acquisition, various efficient strategies have been proposed. In the following discussion, we elaborate on the importance of efficient data storage and retrieval, review some representative acceleration methods for resource allocation and query acceleration, and discuss the challenges associated with them.

\subsubsection{Need for Efficient Data Storage \& Retrieval}
As the amount of data being generated continues to grow exponentially, having a robust and scalable data administration system that can efficiently handle the large data volume and velocity is becoming increasingly critical to support the training of AI models. This need encompasses two aspects. Firstly, data administration systems, such as Hadoop~\cite{hadoop} and Spark~\cite{ZahariaXinEtAl16cacm}, often need to store and merge data from various sources, requiring careful management of memory and computational resources. Secondly, it is crucial to design querying strategies that enable fast data acquisition to ensure timely and accurate processing of the data.

\subsubsection{Resource Allocation} Resource allocation aims to estimate and balance the cost of operations within a data administration system. Two key efficiency metrics in data administration systems are throughput, which refers to how quickly new data can be collected, and latency, which measures how quickly the system can respond to a request. To optimize these metrics, various parameter-tuning techniques have been proposed, including controlling database configuration settings (e.g., buffer pool size) and runtime operations (e.g., percentage of CPU usage and multi-programming level)~\cite{duan2009tuning}. Early tuning methods rely on rules that are based on intuition, experience, data domain knowledge, and industry best practices from sources such as Apache~\cite{apache} and Cloudera~\cite{yarn}. For instance, Hadoop guidelines~\cite{white2012hadoop} suggest that the number of reduced tasks should be set to approximately 0.95 or 1.75 times the number of reduced slots available in the cluster to ensure system tolerance for re-executing failed or slow tasks.

Various learning-based strategies have been developed for resource allocation in data processing systems. For instance, Starfish~\cite{herodotou2011starfish} proposes a profile-predict-optimize approach that generates job profiles with dataflow and cost statistics, which are then used to predict virtual job profiles for task scheduling. More recently, machine learning approaches such as OtterTune~\cite{van2017automatic} have been developed to automatically select the most important parameters, map workloads, and recommend parameters to improve latency and throughput. These learning-based automation strategies can adaptively balance system resources without assuming any internal system information.

\subsubsection{Query Acceleration}
Another research direction is efficient data retrieval, which can be achieved through efficient index selection and query rewriting strategies.

\emph{Query index selection.} The objective of index selection is to minimize the number of disk accesses needed during query processing. To achieve this, programmatic automation strategies create an indexing scheme with indexable columns and record query execution costs~\cite{sun2019end}. Then, they apply either a greedy algorithm~\cite{chaudhuri1997efficient} or dynamic programming~\cite{valentin2000db2} to select the indexing strategy. To enable a more adaptive and flexible querying strategy, learning-based automation strategies collect indexing data from human experts and train machine learning models to predict the proper indexing strategies~\cite{pedrozo2018adaptive}, or search for the optimal strategies using reinforcement learning~\cite{sadri2020online}.

\emph{Query rewriting.} In parallel, query rewriting aims to reduce the workload by identifying repeated sub-queries from input queries. Rule-based strategies~\cite{baik2019bridging,chavan2011dbridge} rewrite queries with pre-defined rules, such as DBridge~\cite{chavan2011dbridge}, which constructs a dependency graph to model the data flow and iteratively applies transformation rules. Learning-based approaches use supervised learning~\cite{he2016learning} or reinforcement learning~\cite{zhou2021dbmind} to predict rewriting rules given an input query.

\subsubsection{Challenges}
Existing data storage and retrieval methods typically focus on optimizing specific parts of the system, such as resource allocation and query acceleration we mentioned. However, the real data administration system as a whole can be complex since it needs to process a vast amount of data in various formats and structures, making end-to-end optimization a challenging task. Additionally, apart from efficiency, data storage and retrieval require consideration of several other crucial and challenging aspects, such as data access control and system maintenance.

%% file: sec6.tex
\section{Data Benchmark}
\label{sec:6}

In the previous sections, we explored a diverse range of data-centric AI tasks throughout various stages of the data lifecycle. Examining benchmarks is a promising approach for gaining insight into the progress of research and development in these tasks, as benchmarks comprehensively evaluate various methods based on standard and agreed-upon metrics. It is important to note that, within the context of data-centric AI, we are specifically interested in \emph{data benchmarks} rather than model benchmarks, which should assess various techniques aimed at achieving data excellence. In this section, we survey the existing benchmarks for different goals of data-centric AI. Firstly, we will introduce the benchmark collection strategy, and subsequently, we will summarize and analyze the collected benchmarks.

\begin{table}[t]
\tiny
\centering
\caption{Data benchmarks. Note that they evaluate \emph{data} rather than model.}
\label{tbl:benchsummary}
\setlength{\tabcolsep}{2.0pt}
\begin{tabular}{cccccc} \toprule
\textbf{Reference} & \textbf{Sub-goal} & \textbf{Task} & \textbf{Domain} & \textbf{Data modality}  & \textbf{Open-source} \\
\midrule
\rowcolor{Gray} 
\multicolumn{6}{c}{\textbf{\textit{Training data development}}} \\ 
Cohen et al.~\cite{cohen2017publicly} & Collection & Dataset discovery & Biomedical & Tabular, text & \xmark \\
Poess et al.~\cite{poess2014tpc} & Collection & Data integration & Database & Tabular, time-series & \xmark \\
Pinkel et al.~\cite{pinkel2015rodi} & Collection & Data integration & Database & Tabular, graph & \xmark \\
Wang et al.~\cite{wang2022usb} & Labeling & Semi-supervised learning & AI & Image, text, audio & \cmark \\
Yang et al.~\cite{yang2018benchmark} & Labeling & Active learning & AI & Tabular, image, text & \xmark \\
Meduri et al.~\cite{meduri2020comprehensive} & Labeling & Active learning & Database & Tabular, text & \xmark \\
Abdelaal et al.~\cite{abdelaal2023rein} & Preparation & Data cleaning & Database & Tabular, text, time-series & \cmark \\
Li et al.~\cite{li2019cleanml} & Preparation & Data cleaning & Database & Tabular, time-series & \cmark \\
J{\"a}ger et al.~\cite{jager2021benchmark} &  Preparation & Data cleaning & AI & Tabular, image & \xmark \\ 
Buckley et al.~\cite{buckley2022feature} & Preparation & Feature extraction & Healthcare & Tabular, image, time-series & \cmark \\ 
Vijayan et al.~\cite{vijayan2022blood} & Preparation & Feature extraction & Biomedical & Tabular, sequential & \cmark \\
Bommert et al.~\cite{bommert2022benchmark} & Reduction & Feature selection & Biomedical & Tabular, sequential & \cmark \\
Espadoto et al.~\cite{espadoto2019toward} & Reduction & Dimensionality reduction & Computer graphics & Tabular, image, audio & \cmark \\
Grochowski et al.~\cite{grochowski2004comparison} & Reduction & Instance selection & Computer graphics & Tabular, image, audio & \cmark \\
Blachnik et al.~\cite{blachnik2020comparison} & Reduction & Instance selection & Computer graphics & Tabular, image, audio & \cmark \\
Iwana et al.~\cite{iwana2021empirical} & Augmentation & All sub-goals & AI & Time-series & \cmark \\
Nanni et al.~\cite{nanni2021comparison} & Augmentation & Basic manipulation & AI & Image & \cmark \\
Yoo et al.~\cite{yoo2020rethinking} & Augmentation & Basic manipulation & AI & Image & \cmark \\
Ding et al.~\cite{ding2022data} & Augmentation & Augmentation data synthesis & AI & Graph & \xmark \\
Tao et al.~\cite{tao2021benchmarking} & Augmentation & Augmentation data synthesis & Computer security & Tabular & \xmark \\
Zoller et al.~\cite{zoller2021benchmark} & - & Pipeline search & AI & Tabular, image, audio, time-series & \cmark \\
Gijsbers et al.~\cite{gijsbers2022amlb} & - & Pipeline search & AI & Tabular, image, audio, time-series & \cmark \\
\rowcolor{Gray} 
\multicolumn{6}{c}{\textbf{\textit{Evaluation data development}}} \\ 
Srivastava et al.~\cite{srivastava2022beyond} & In-distribution & Evaluation data synthesis & AI & Text & \cmark \\
Pawelczyk et al.~\cite{pawelczyk2021carla} & In-distribution & Algorithmic recourse & AI & Tabular & \cmark \\
Dong et al.~\cite{dong2020benchmarking} & Out-of-distribution & Adversarial samples & AI & Image & \cmark \\
Hendrycks et al.~\cite{hendrycks2019benchmarking} & Out-of-distribution & Adversarial samples & AI & Image & \cmark \\
Yoo et al.~\cite{yoo2020searching} & Out-of-distribution & Adversarial samples  & AI & Text & \cmark \\
\rowcolor{Gray} 
\multicolumn{6}{c}{\textbf{\textit{Data maintenance}}} \\ 
Kanthara et al.~\cite{kanthara2022chart} & Understanding & Visual summarization  & AI & Tabular, text & \cmark \\
Grinstein et al.~\cite{grinstein2002benchmark} & Understanding & Visual summarization & Human-computer interaction & Tabular, image & \cmark \\
Zeng et al.~\cite{zeng2021evaluation} & Understanding & Visualization recommendation  & Human-computer Interaction & Tabular & \xmark \\
Jia et al.~\cite{jia2021scalability} & Understanding & Data valuation & AI & Image & \cmark \\
Batini et al.~\cite{batini2009methodologies} & Quality assurance & Quality assessment  & Database & Tabular & \xmark \\
Arocena et al.~\cite{arocena2016benchmarking} & Quality assurance & Quality improvement  & Database & Tabular & \xmark \\
Zhang et al.~\cite{zhang2022facilitating} & Storage \& retrieval & Resource allocation  & Database & Tabular & \cmark \\
Marcus et al.~\cite{marcus2020benchmarking} & Storage \& retrieval & Query index selection  & Database & Tabular & \xmark \\
\rowcolor{Gray} 
\multicolumn{6}{c}{\textbf{\textit{Unified benchmark}}} \\ 
Mazumder et al.~\cite{mazumder2022dataperf} & Multiple & 6 distinct tasks & AI & Multiple & \xmark \\
\bottomrule
\end{tabular}
\end{table}

\emph{Collection strategy.} We primarily utilize Google Scholar to search for benchmark papers. Specifically, we generate a series of queries for each task using relevant keywords for the sub-goal and task, and supplement them with terms such as ``benchmark'', ``quantitative analysis'', and ``quantitative survey''. For example, the queries for the task ``data cleaning'' include ``benchmark data cleaning'', ``benchmark data cleansing'', ``quantitative analysis for data cleaning'', ``quantitative survey for data cleaning'', etc. It is worth noting that many of the queried benchmarks evaluate models rather than data. Thus, we have carefully read each paper and manually filtered the papers to ensure that they focus on the evaluation of data. We have also screened them based on the number of citations and the reputation of the publication venues.

\emph{Summary of the collected benchmarks.} Table~\ref{tbl:benchsummary} comprises the 36 benchmarks that we collected using the above process, out of which 23 incorporate open-source codes. Notably, we did not encounter a benchmark for the task of ``generating distribution shift samples'', although there are benchmarks available for \emph{detecting} distribution-shifted samples~\cite{koh2021wilds}. We omitted it from the table since it mainly assesses model performance on distribution shift rather than discussing how to create distribution-shifted data that can expose model weaknesses.

\emph{Meta-analysis.} We give a bird-eye view of existing data-centric AI research across various dimensions by analyzing these collected benchmarks. \textbf{\ding{182}} Although the AI community has made the most significant contributions to these benchmarks (17), numerous other domains have also made substantial contributions, including databases (9), computer graphics (3), human-computer interaction (2), biomedical (3), computer security (1), and healthcare (1). Notably, healthcare and biomedical are outside the realm of computer science. An established benchmark in a domain often implies that there is a collection of published works. Therefore, data-centric AI is an interdisciplinary effort that spans various domains within and outside of computer science. \textbf{\ding{183}} The most frequently benchmarked data modality is tabular data (25), followed by image (15), time-series (7), text (6), audio (6), and graph (2). We conjecture that this is because tabular and image data have been extensively studied, while research on graph data is still emerging. \textbf{\ding{184}} Training data development has received more attention, if we measure it based on the number of benchmarks (22), compared to evaluation data development (5) and data maintenance (8). We hypothesize that this is due to the fact that many of the tasks involved in training data development were considered as preprocessing steps in the model-centric paradigm.





%% file: sec7.tex
\section{Discussion and Future Direction}
\label{sec:7}

What is the current stage of data-centric AI research, and what are the potential future directions? This section provides a top-level discussion of data-centric AI and presents some of the open problems that we have identified, aiming to motivate future exploration in this field. We start by trying to answer the research questions posed at the beginning:

\begin{itemize}
    \item \emph{RQ1: What are the necessary tasks to make AI data-centric?} Data-centric AI encompasses a range of tasks that involve developing training data, inference data, and maintaining data. These tasks include but are not limited to 1) cleaning, labeling, preparing, reducing, and augmenting the training data, 2) generating in-distribution and out-of-distribution data for evaluation, or tuning prompts to achieve desired outcomes, and 3) constructing efficient infrastructures for understanding, organizing, and debugging data.
    \item \emph{RQ2: Why is automation significant for developing and maintaining data?} Given the availability of an increasing amount of data at an unprecedented rate, it is imperative to develop automated algorithms to streamline the process of data development and maintenance. Based on the papers surveyed in Tables~\ref{tbl:trainingmethodsummary}, \ref{tbl:inferencemethodsummary}, and \ref{tbl:maintainmethodsummary}, automated algorithms have been developed for all sub-goals. These automation algorithms span different automation levels, from programmatic automation to learning-based automation, to pipeline automation.
    \item \emph{RQ3: In which cases and why is human participation essential in data-centric AI?} Human participation is necessary for many data-centric AI tasks, such as the majority of data labeling tasks (Table~\ref{tbl:trainingmethodsummary}) and several tasks in inference data development (Table~\ref{tbl:inferencemethodsummary}). Notably, different methods may require varying degrees of human participation, ranging from full involvement to providing minimal inputs. Human participation is crucial in many scenarios because it is often the only way to ensure that the behavior of AI systems aligns with human intentions.
    \item \emph{RQ4: What is the current progress of data-centric AI?} Although data-centric AI is a relatively new concept, considerable progress has already been made in many relevant tasks, the majority of which were viewed as preprocessing steps in the model-centric paradigm. Meanwhile, many new tasks have recently emerged, and research on them is still ongoing. In Section~\ref{sec:6}, our meta-analysis on benchmark papers reveals that progress has been made across different domains, with the majority of the benchmarks coming from the AI domain. Among the three general data-centric AI goals, training data development has received relatively more research attention. For data modality, tabular and image data have been the primary focus. As research papers on data-centric AI are growing exponentially~\cite{zha2023data}, we could witness even more progress in this field in the future.
\end{itemize}
By attempting to address these questions, our survey delves into a variety of tasks and their needs and challenges, yielding a more concrete picture of the scope and progress of data-centric AI. However, although we have endeavored to broadly and comprehensively cover various tasks and techniques, it is impossible to include every aspect of data-centric AI. In the following, we connect data-centric AI with two other popular research topics in AI:

\begin{itemize}
    \item \emph{Foundation models.} A foundation model is a large model that is trained on massive amounts of unlabeled data and can be adapted to various tasks, such as large language models~\cite{brown2020language,gpt4}, and Stable Diffusion~\cite{rombach2022high}. As models become sufficiently powerful, it becomes feasible to perform many data-centric AI tasks with models, such as data labeling~\cite{gpt4}, and data augmentation~\cite{yoo2021gpt3mix}. Consequently, the recent trend of foundation models has the potential to fundamentally alter our understanding of data. Unlike the conventional approach of storing raw data values in datasets, the model itself can be a form of data (or a ``container'' of raw data) since the model can convey information (see the definition of data in Section~\ref{sec:2:1}). Foundation models blur the boundary between data and model, but their training still heavily relies on large and high-quality datasets.
    \item \emph{Reinforcement learning.} Reinforcement learning is a research field that trains intelligent agents to optimize rewards without any initial data~\cite{mnih2013playing,zha2021douzero,zha2021rank,zha2021rlcard,zha2019experience,zha2022autoshard,zhadreamshard,lai2020dual,zha2021simplifying}. It is a unique learning paradigm that alternates between generating data with the model and training the model with self-generated data. Like foundation models, the advancement of reinforcement learning could also possibly blur the boundary between data and model. Furthermore, reinforcement learning has already been widely adopted in several data-centric AI sub-goals, such as data labeling~\cite{christiano2017deep,dong2023active,zha2020meta}, data preparation~\cite{khurana2018feature}, data reduction~\cite{liu2020mesa}, and data augmentation~\cite{cubuk2019autoaugment,zha2022towards}. The reason could be attributed to its goal-oriented nature, which is well-suited for automation.
\end{itemize}

Upon examining the connections to these two rapidly evolving research fields, we hypothesize that data-centric AI and model-centric AI could become even more intertwined in the development of AI systems. Looking forward, we present some potential future directions we have identified in data-centric AI:
\begin{itemize}
    \item \emph{Cross-task automation.} While there has been significant progress in automating various individual data-centric AI tasks, joint automation across multiple tasks remains largely unexplored. Although pipeline search methods~\cite{feurer2015efficient,lai2021tods,zha2021autovideo,heffetz2020deepline} have emerged, they are limited only to training data development. From a broad data-centric AI perspective, it would be desirable to have a unified framework for jointly automating tasks aimed at different goals, ranging from training data development to inference data development and data maintenance.
    \item \emph{Data-model co-design.} Although data-centric AI advocates for shifting the focus to data, it does not necessarily imply that the model has to remain unchanged. The optimal data strategies may differ when using different models, and vice versa. Furthermore, as discussed above, the boundary between data and model could potentially become increasingly blurred with the advancement of foundation models and reinforcement learning. Consequently, future progress in AI could arise from co-designing data and models, and the co-evolution of data and models could pave the way toward more powerful AI systems.
    \item \emph{Debiasing data.} In many high-stakes applications, AI systems have recently been found to exhibit discriminatory behavior towards certain groups of people, sparking significant concerns about fairness~\cite{mehrabi2021survey,wan2022processing,ding2023fairly,chuang2022mitigating,jiang2022generalized,jiang2022fmp}. These biases often originate from imbalanced distributions of sensitive variables in the data. From a data-centric perspective, more research efforts are needed to debias data, including but limited to mitigating biases in training data, systematic methodologies to construct evaluation data to expose unfairness issues of unfairness, and continuously maintaining fair data in a dynamic environment.
    \item \emph{Tackling data in various modalities.} Based on the benchmark analysis presented in Section~\ref{sec:6}, most research efforts have been directed toward tabular and image data. However, other data modalities that are comparably important but less studied in data-centric AI pose significant challenges. For instance, time-series data~\cite{zha2022towards2,hamilton2020time,li2022towards} exhibit complex temporal correlations, while graph data~\cite{zhou2020graph,zhou2021multi,zhou2020towards,lai2020policy,zhou2021dirichlet,tan2023bring,liu2022rsc} has intricate data dependencies. Therefore, more research on how to engineer data for these modalities is required. Furthermore, developing data-centric AI solutions that can simultaneously address multiple data modalities is an intriguing avenue for future exploration.
    \item \emph{Data benchmarks development.} The advancement of model-centric AI has been facilitated by benchmarks in advancing model designs. Whereas data-centric AI requires more attention to benchmarking. As discussed in Section~\ref{sec:6}, existing benchmarks for data-centric AI typically only focus on specific tasks. Constructing a unified benchmark to evaluate overall data quality and various data-centric AI techniques comprehensively presents a significant challenge. Although DataPerf~\cite{mazumder2022dataperf} has made notable progress towards this objective, it currently supports only a limited number of tasks. The development of more unified data benchmarks would greatly accelerate research progress in this area.
\end{itemize}

%% file: sec8.tex
\section{Conclusion}
\label{sec:8}

This survey focuses on data-centric AI, an emerging and important research field in AI. We motivated the need for data-centric AI by showing how carefully designing and maintaining data can make AI solutions more desirable across academia and industry. Next, we provided a background of data-centric AI, which includes its definition and a goal-driven taxonomy. Then, guided by the research questions posed, we reviewed various data-centric AI techniques for different purposes from the perspectives of automation and collaboration. Furthermore, we collected data benchmarks from different domains and analyzed them at a meta-level. Lastly, we discussed data-centric AI from a global view and shared our perspectives on the blurred boundaries between data and model. We also presented potential future directions for this field. To conclude in one line, we believe that data will play an increasingly important role in building AI systems. At the same time, there are still numerous challenges that need to be addressed.  We hope our survey could inspire collaborative initiatives in our community to push forward this field.

%% file: main.bbl
\begin{thebibliography}{100}

\bibitem{abdelaal2023rein}
{\sc Abdelaal, M., Hammacher, C., and Schoening, H.}
\newblock Rein: A comprehensive benchmark framework for data cleaning methods
  in ml pipelines.
\newblock {\em arXiv preprint arXiv:2302.04702\/} (2023).

\bibitem{abdi2010principal}
{\sc Abdi, H., and Williams, L.~J.}
\newblock Principal component analysis.
\newblock {\em Wiley interdisciplinary reviews: computational statistics 2}, 4
  (2010), 433--459.

\bibitem{agarwal2019marketplace}
{\sc Agarwal, A., Dahleh, M., and Sarkar, T.}
\newblock A marketplace for data: An algorithmic solution.
\newblock In {\em EC\/} (2019).

\bibitem{ahsan2021effect}
{\sc Ahsan, M.~M., Mahmud, M.~P., Saha, P.~K., Gupta, K.~D., and Siddique, Z.}
\newblock Effect of data scaling methods on machine learning algorithms and
  model performance.
\newblock {\em Technologies 9}, 3 (2021), 52.

\bibitem{ali2014data}
{\sc Ali, P. J.~M., Faraj, R.~H., Koya, E., Ali, P. J.~M., and Faraj, R.~H.}
\newblock Data normalization and standardization: a technical report.
\newblock {\em Mach Learn Tech Rep 1}, 1 (2014), 1--6.

\bibitem{apache}
{\sc Apache}.
\newblock Apache.
\newblock {\em https://storm.apache.org/releases/current/Performance.html\/}
  (2023).

\bibitem{armbrust2021lakehouse}
{\sc Armbrust, M., Ghodsi, A., Xin, R., and Zaharia, M.}
\newblock Lakehouse: a new generation of open platforms that unify data
  warehousing and advanced analytics.
\newblock In {\em CIDR\/} (2021).

\bibitem{arocena2016benchmarking}
{\sc Arocena, P.~C., Glavic, B., Mecca, G., Miller, R.~J., Papotti, P., and
  Santoro, D.}
\newblock Benchmarking data curation systems.
\newblock {\em IEEE Data Eng. Bull. 39}, 2 (2016), 47--62.

\bibitem{aroyo2022data}
{\sc Aroyo, L., Lease, M., Paritosh, P., and Schaekermann, M.}
\newblock Data excellence for ai: why should you care?
\newblock {\em Interactions 29}, 2 (2022), 66--69.

\bibitem{azhagusundari2013feature}
{\sc Azhagusundari, B., Thanamani, A.~S., et~al.}
\newblock Feature selection based on information gain.
\newblock {\em International Journal of Innovative Technology and Exploring
  Engineering (IJITEE) 2}, 2 (2013), 18--21.

\bibitem{azizzadenesheli2019regularized}
{\sc Azizzadenesheli, K., Liu, A., Yang, F., and Anandkumar, A.}
\newblock Regularized learning for domain adaptation under label shifts.
\newblock {\em arXiv preprint arXiv:1903.09734\/} (2019).

\bibitem{baik2019bridging}
{\sc Baik, C., Jagadish, H.~V., and Li, Y.}
\newblock Bridging the semantic gap with sql query logs in natural language
  interfaces to databases.
\newblock In {\em ICDE\/} (2019).

\bibitem{bank2020autoencoders}
{\sc Bank, D., Koenigstein, N., and Giryes, R.}
\newblock Autoencoders.
\newblock {\em arXiv preprint arXiv:2003.05991\/} (2020).

\bibitem{barandas2020tsfel}
{\sc Barandas, M., Folgado, D., Fernandes, L., Santos, S., Abreu, M., Bota, P.,
  Liu, H., Schultz, T., and Gamboa, H.}
\newblock Tsfel: Time series feature extraction library.
\newblock {\em SoftwareX 11\/} (2020), 100456.

\bibitem{barclay2000microsoft}
{\sc Barclay, T., Gray, J., and Slutz, D.}
\newblock Microsoft terraserver: a spatial data warehouse.
\newblock In {\em SIGMOD\/} (2000).

\bibitem{barenstein2019propublica}
{\sc Barenstein, M.}
\newblock Propublica's compas data revisited.
\newblock {\em arXiv preprint arXiv:1906.04711\/} (2019).

\bibitem{basu1995discovering}
{\sc Basu, A., and Blanning, R.~W.}
\newblock Discovering implicit integrity constraints in rule bases using
  metagraphs.
\newblock In {\em HICSS\/} (1995).

\bibitem{batini2009methodologies}
{\sc Batini, C., Cappiello, C., Francalanci, C., and Maurino, A.}
\newblock Methodologies for data quality assessment and improvement.
\newblock {\em ACM computing surveys (CSUR) 41}, 3 (2009), 1--52.

\bibitem{baylor2017tfx}
{\sc Baylor, D., Breck, E., Cheng, H.-T., Fiedel, N., Foo, C.~Y., Haque, Z.,
  Haykal, S., Ispir, M., Jain, V., Koc, L., et~al.}
\newblock Tfx: A tensorflow-based production-scale machine learning platform.
\newblock In {\em KDD\/} (2017).

\bibitem{becker2021step}
{\sc Becker, M., Burkart, N., Birnstill, P., and Beyerer, J.}
\newblock A step towards global counterfactual explanations: Approximating the
  feature space through hierarchical division and graph search.
\newblock {\em Adv. Artif. Intell. Mach. Learn. 1}, 2 (2021), 90--110.

\bibitem{betzalel2022study}
{\sc Betzalel, E., Penso, C., Navon, A., and Fetaya, E.}
\newblock A study on the evaluation of generative models.
\newblock {\em arXiv preprint arXiv:2206.10935\/} (2022).

\bibitem{bhardwaj2015datahub}
{\sc Bhardwaj, A., Bhattacherjee, S., Chavan, A., Deshpande, A., Elmore, A.~J.,
  Madden, S., and Parameswaran, A.~G.}
\newblock Datahub: Collaborative data science \& dataset version management at
  scale.
\newblock In {\em CIDR\/} (2015).

\bibitem{biggio2013evasion}
{\sc Biggio, B., Corona, I., Maiorca, D., Nelson, B., {\v{S}}rndi{\'c}, N.,
  Laskov, P., Giacinto, G., and Roli, F.}
\newblock Evasion attacks against machine learning at test time.
\newblock In {\em ECMLPKDD\/} (2013).

\bibitem{bisong2019introduction}
{\sc Bisong, E., and Bisong, E.}
\newblock Introduction to scikit-learn.
\newblock {\em Building Machine Learning and Deep Learning Models on Google
  Cloud Platform: A Comprehensive Guide for Beginners\/} (2019), 215--229.

\bibitem{blachnik2020comparison}
{\sc Blachnik, M., and Kordos, M.}
\newblock Comparison of instance selection and construction methods with
  various classifiers.
\newblock {\em Applied Sciences 10}, 11 (2020), 3933.

\bibitem{blanchart2021exact}
{\sc Blanchart, P.}
\newblock An exact counterfactual-example-based approach to tree-ensemble
  models interpretability.
\newblock {\em arXiv preprint arXiv:2105.14820\/} (2021).

\bibitem{boeckinginteractive}
{\sc Boecking, B., Neiswanger, W., Xing, E., and Dubrawski, A.}
\newblock Interactive weak supervision: Learning useful heuristics for data
  labeling.
\newblock In {\em ICLR\/} (2021).

\bibitem{bogatu2020dataset}
{\sc Bogatu, A., Fernandes, A.~A., Paton, N.~W., and Konstantinou, N.}
\newblock Dataset discovery in data lakes.
\newblock In {\em ICDE\/} (2020).

\bibitem{bohannon2006conditional}
{\sc Bohannon, P., Fan, W., Geerts, F., Jia, X., and Kementsietsidis, A.}
\newblock Conditional functional dependencies for data cleaning.
\newblock In {\em 2007 IEEE 23rd international conference on data
  engineering\/} (2006), IEEE, pp.~746--755.

\bibitem{bollacker2008freebase}
{\sc Bollacker, K., Evans, C., Paritosh, P., Sturge, T., and Taylor, J.}
\newblock Freebase: a collaboratively created graph database for structuring
  human knowledge.
\newblock In {\em SIGMOD\/} (2008).

\bibitem{bommert2022benchmark}
{\sc Bommert, A., Welchowski, T., Schmid, M., and Rahnenf{\"u}hrer, J.}
\newblock Benchmark of filter methods for feature selection in high-dimensional
  gene expression survival data.
\newblock {\em Briefings in Bioinformatics 23}, 1 (2022), bbab354.

\bibitem{borgwardt2006integrating}
{\sc Borgwardt, K.~M., Gretton, A., Rasch, M.~J., Kriegel, H.-P.,
  Sch{\"o}lkopf, B., and Smola, A.~J.}
\newblock Integrating structured biological data by kernel maximum mean
  discrepancy.
\newblock {\em Bioinformatics 22}, 14 (2006), e49--e57.

\bibitem{borkin2013makes}
{\sc Borkin, M.~A., Vo, A.~A., Bylinskii, Z., Isola, P., Sunkavalli, S., Oliva,
  A., and Pfister, H.}
\newblock What makes a visualization memorable?
\newblock {\em IEEE transactions on visualization and computer graphics 19}, 12
  (2013), 2306--2315.

\bibitem{brown2020language}
{\sc Brown, T., Mann, B., Ryder, N., Subbiah, M., Kaplan, J.~D., Dhariwal, P.,
  Neelakantan, A., Shyam, P., Sastry, G., Askell, A., et~al.}
\newblock Language models are few-shot learners.
\newblock {\em NeurIPS\/} (2020).

\bibitem{buckley2022feature}
{\sc Buckley, T., Ghosh, B., and Pakrashi, V.}
\newblock A feature extraction \& selection benchmark for structural health
  monitoring.
\newblock {\em Structural Health Monitoring\/} (2022), 14759217221111141.

\bibitem{buolamwini2018gender}
{\sc Buolamwini, J., and Gebru, T.}
\newblock Gender shades: Intersectional accuracy disparities in commercial
  gender classification.
\newblock In {\em FAccT\/} (2018).

\bibitem{burch2014benefits}
{\sc Burch, M., and Weiskopf, D.}
\newblock On the benefits and drawbacks of radial diagrams.
\newblock {\em Handbook of human centric visualization\/} (2014), 429--451.

\bibitem{carreira2021counterfactual}
{\sc Carreira-Perpin{\'a}n, M.~A., and Hada, S.~S.}
\newblock Counterfactual explanations for oblique decision trees: Exact,
  efficient algorithms.
\newblock In {\em AAAI\/} (2021).

\bibitem{chaudhuri1997efficient}
{\sc Chaudhuri, S., and Narasayya, V.~R.}
\newblock An efficient, cost-driven index selection tool for microsoft sql
  server.
\newblock In {\em VLDB\/} (1997).

\bibitem{chavan2011dbridge}
{\sc Chavan, M., Guravannavar, R., Ramachandra, K., and Sudarshan, S.}
\newblock Dbridge: A program rewrite tool for set-oriented query execution.
\newblock In {\em ICDE\/} (2011).

\bibitem{chawla2002smote}
{\sc Chawla, N.~V., Bowyer, K.~W., Hall, L.~O., and Kegelmeyer, W.~P.}
\newblock Smote: synthetic minority over-sampling technique.
\newblock {\em Journal of artificial intelligence research 16\/} (2002),
  321--357.

\bibitem{chen2020mixtext}
{\sc Chen, J., Yang, Z., and Yang, D.}
\newblock Mixtext: Linguistically-informed interpolation of hidden space for
  semi-supervised text classification.
\newblock In {\em ACL\/} (2020).

\bibitem{chen2017zoo}
{\sc Chen, P.-Y., Zhang, H., Sharma, Y., Yi, J., and Hsieh, C.-J.}
\newblock Zoo: Zeroth order optimization based black-box attacks to deep neural
  networks without training substitute models.
\newblock In {\em AISec Workshop\/} (2017).

\bibitem{chen2020building}
{\sc Chen, T., Han, L., Demartini, G., Indulska, M., and Sadiq, S.}
\newblock Building data curation processes with crowd intelligence.
\newblock In {\em CAiSE\/} (2020).

\bibitem{chlap2021review}
{\sc Chlap, P., Min, H., Vandenberg, N., Dowling, J., Holloway, L., and
  Haworth, A.}
\newblock A review of medical image data augmentation techniques for deep
  learning applications.
\newblock {\em Journal of Medical Imaging and Radiation Oncology 65}, 5 (2021),
  545--563.

\bibitem{chong2020graph}
{\sc Chong, Y., Ding, Y., Yan, Q., and Pan, S.}
\newblock Graph-based semi-supervised learning: A review.
\newblock {\em Neurocomputing 408\/} (2020), 216--230.

\bibitem{chowdhary2020natural}
{\sc Chowdhary, K., and Chowdhary, K.}
\newblock Natural language processing.
\newblock {\em Fundamentals of artificial intelligence\/} (2020), 603--649.

\bibitem{christiano2017deep}
{\sc Christiano, P.~F., Leike, J., Brown, T., Martic, M., Legg, S., and Amodei,
  D.}
\newblock Deep reinforcement learning from human preferences.
\newblock In {\em NeurIPS\/} (2017).

\bibitem{chu2013discovering}
{\sc Chu, X., Ilyas, I.~F., and Papotti, P.}
\newblock Discovering denial constraints.
\newblock In {\em VLDB\/} (2013).

\bibitem{chuang2022mitigating}
{\sc Chuang, Y.-N., Lai, K.-H., Tang, R., Du, M., Chang, C.-Y., Zou, N., and
  Hu, X.}
\newblock Mitigating relational bias on knowledge graphs.
\newblock {\em arXiv preprint arXiv:2211.14489\/} (2022).

\bibitem{chuang2023efficient}
{\sc Chuang, Y.-N., Wang, G., Yang, F., Liu, Z., Cai, X., Du, M., and Hu, X.}
\newblock Efficient xai techniques: A taxonomic survey.
\newblock {\em arXiv preprint arXiv:2302.03225\/} (2023).

\bibitem{chuangcortx}
{\sc Chuang, Y.-N., Wang, G., Yang, F., Zhou, Q., Tripathi, P., Cai, X., and
  Hu, X.}
\newblock Cortx: Contrastive framework for real-time explanation.
\newblock In {\em ICLR\/} (2023).

\bibitem{chung2019slice}
{\sc Chung, Y., Kraska, T., Polyzotis, N., Tae, K.~H., and Whang, S.~E.}
\newblock Slice finder: Automated data slicing for model validation.
\newblock In {\em ICDE\/} (2019).

\bibitem{cohen2017publicly}
{\sc Cohen, T., Roberts, K., Gururaj, A.~E., Chen, X., Pournejati, S., Alter,
  G., Hersh, W.~R., Demner-Fushman, D., Ohno-Machado, L., and Xu, H.}
\newblock A publicly available benchmark for biomedical dataset retrieval: the
  reference standard for the 2016 biocaddie dataset retrieval challenge.
\newblock {\em Database 2017\/} (2017).

\bibitem{cohn1996active}
{\sc Cohn, D.~A., Ghahramani, Z., and Jordan, M.~I.}
\newblock Active learning with statistical models.
\newblock {\em Journal of artificial intelligence research 4\/} (1996),
  129--145.

\bibitem{cubuk2019autoaugment}
{\sc Cubuk, E.~D., Zoph, B., Mane, D., Vasudevan, V., and Le, Q.~V.}
\newblock Autoaugment: Learning augmentation policies from data.
\newblock In {\em CVPR\/} (2019).

\bibitem{dandl2020multi}
{\sc Dandl, S., Molnar, C., Binder, M., and Bischl, B.}
\newblock Multi-objective counterfactual explanations.
\newblock In {\em PPSN\/} (2020).

\bibitem{dekel2009vox}
{\sc Dekel, O., and Shamir, O.}
\newblock Vox populi: Collecting high-quality labels from a crowd.
\newblock In {\em COLT\/} (2009).

\bibitem{deng2009imagenet}
{\sc Deng, J., Dong, W., Socher, R., Li, L.-J., Li, K., and Fei-Fei, L.}
\newblock Imagenet: A large-scale hierarchical image database.
\newblock In {\em CVPR\/} (2009).

\bibitem{deodhar2022human}
{\sc Deodhar, M., Ma, X., Cai, Y., Koes, A., Beutel, A., and Chen, J.}
\newblock A human-ml collaboration framework for improving video content
  reviews.
\newblock {\em arXiv preprint arXiv:2210.09500\/} (2022).

\bibitem{desnoyers2011toward}
{\sc Desnoyers, L.}
\newblock Toward a taxonomy of visuals in science communication.
\newblock {\em Technical Communication 58}, 2 (2011), 119--134.

\bibitem{dhurandhar2019model}
{\sc Dhurandhar, A., Pedapati, T., Balakrishnan, A., Chen, P.-Y., Shanmugam,
  K., and Puri, R.}
\newblock Model agnostic contrastive explanations for structured data.
\newblock {\em arXiv preprint arXiv:1906.00117\/} (2019).

\bibitem{ding2021retiring}
{\sc Ding, F., Hardt, M., Miller, J., and Schmidt, L.}
\newblock Retiring adult: New datasets for fair machine learning.
\newblock In {\em NeurIPS\/} (2021).

\bibitem{ding2022data}
{\sc Ding, K., Xu, Z., Tong, H., and Liu, H.}
\newblock Data augmentation for deep graph learning: A survey.
\newblock {\em ACM SIGKDD Explorations Newsletter 24}, 2 (2022), 61--77.

\bibitem{ding2023fairly}
{\sc Ding, S., Tang, R., Zha, D., Zou, N., Zhang, K., Jiang, X., and Hu, X.}
\newblock Fairly predicting graft failure in liver transplant for organ
  assigning.
\newblock {\em arXiv preprint arXiv:2302.09400\/} (2023).

\bibitem{dong2023active}
{\sc Dong, J., Zhang, Q., Huang, X., Tan, Q., Zha, D., and Zhao, Z.}
\newblock Active ensemble learning for knowledge graph error detection.
\newblock In {\em WSDM\/} (2023).

\bibitem{dong2020benchmarking}
{\sc Dong, Y., Fu, Q.-A., Yang, X., Pang, T., Su, H., Xiao, Z., and Zhu, J.}
\newblock Benchmarking adversarial robustness on image classification.
\newblock In {\em CVPR\/} (2020).

\bibitem{drori2021alphad3m}
{\sc Drori, I., Krishnamurthy, Y., Rampin, R., Lourenco, R. d.~P., Ono, J.~P.,
  Cho, K., Silva, C., and Freire, J.}
\newblock Alphad3m: Machine learning pipeline synthesis.
\newblock {\em arXiv preprint arXiv:2111.02508\/} (2021).

\bibitem{duan2009tuning}
{\sc Duan, S., Thummala, V., and Babu, S.}
\newblock Tuning database configuration parameters with ituned.
\newblock In {\em VLDB\/} (2009).

\bibitem{espadoto2019toward}
{\sc Espadoto, M., Martins, R.~M., Kerren, A., Hirata, N.~S., and Telea, A.~C.}
\newblock Toward a quantitative survey of dimension reduction techniques.
\newblock {\em IEEE transactions on visualization and computer graphics 27}, 3
  (2019), 2153--2173.

\bibitem{eykholt2018robust}
{\sc Eykholt, K., Evtimov, I., Fernandes, E., Li, B., Rahmati, A., Xiao, C.,
  Prakash, A., Kohno, T., and Song, D.}
\newblock Robust physical-world attacks on deep learning visual classification.
\newblock In {\em CVPR\/} (2018).

\bibitem{fahad2014survey}
{\sc Fahad, A., Alshatri, N., Tari, Z., Alamri, A., Khalil, I., Zomaya, A.~Y.,
  Foufou, S., and Bouras, A.}
\newblock A survey of clustering algorithms for big data: Taxonomy and
  empirical analysis.
\newblock {\em IEEE transactions on emerging topics in computing 2}, 3 (2014),
  267--279.

\bibitem{farahani2021brief}
{\sc Farahani, A., Voghoei, S., Rasheed, K., and Arabnia, H.~R.}
\newblock A brief review of domain adaptation.
\newblock {\em Advances in Data Science and Information Engineering:
  Proceedings from ICDATA 2020 and IKE 2020\/} (2021), 877--894.

\bibitem{feng2021survey}
{\sc Feng, S.~Y., Gangal, V., Wei, J., Chandar, S., Vosoughi, S., Mitamura, T.,
  and Hovy, E.}
\newblock A survey of data augmentation approaches for nlp.
\newblock In {\em ACL\/} (2021).

\bibitem{fernandez2018aurum}
{\sc Fernandez, R.~C., Abedjan, Z., Koko, F., Yuan, G., Madden, S., and
  Stonebraker, M.}
\newblock Aurum: A data discovery system.
\newblock In {\em ICDE\/} (2018).

\bibitem{feurer2015efficient}
{\sc Feurer, M., Klein, A., Eggensperger, K., Springenberg, J., Blum, M., and
  Hutter, F.}
\newblock Efficient and robust automated machine learning.
\newblock In {\em NeurIPS\/} (2015).

\bibitem{hadoop}
{\sc Foundation, A.~S.}
\newblock Hadoop.
\newblock {\em https://hadoop.apache.org\/} (2023).

\bibitem{franconeri2021science}
{\sc Franconeri, S.~L., Padilla, L.~M., Shah, P., Zacks, J.~M., and Hullman,
  J.}
\newblock The science of visual data communication: What works.
\newblock {\em Psychological Science in the public interest 22}, 3 (2021),
  110--161.

\bibitem{frid2018synthetic}
{\sc Frid-Adar, M., Klang, E., Amitai, M., Goldberger, J., and Greenspan, H.}
\newblock Synthetic data augmentation using gan for improved liver lesion
  classification.
\newblock In {\em ISBI\/} (2018).

\bibitem{galhotra2021adaptive}
{\sc Galhotra, S., Golshan, B., and Tan, W.-C.}
\newblock Adaptive rule discovery for labeling text data.
\newblock In {\em SIGMOD\/} (2021).

\bibitem{gamboa2022human}
{\sc Gamboa, E., Libreros, A., Hirth, M., and Dubiner, D.}
\newblock Human-ai collaboration for improving the identification of cars for
  autonomous driving.
\newblock In {\em CIKM Workshop\/} (2022).

\bibitem{gao2021making}
{\sc Gao, T., Fisch, A., and Chen, D.}
\newblock Making pre-trained language models better few-shot learners.
\newblock In {\em ACL\/} (2021).

\bibitem{ghorbani2020distributional}
{\sc Ghorbani, A., Kim, M., and Zou, J.}
\newblock A distributional framework for data valuation.
\newblock In {\em ICML\/} (2020).

\bibitem{ghorbani2019data}
{\sc Ghorbani, A., and Zou, J.}
\newblock Data shapley: Equitable valuation of data for machine learning.
\newblock In {\em ICML\/} (2019).

\bibitem{gijsbers2022amlb}
{\sc Gijsbers, P., Bueno, M.~L., Coors, S., LeDell, E., Poirier, S., Thomas,
  J., Bischl, B., and Vanschoren, J.}
\newblock Amlb: an automl benchmark.
\newblock {\em arXiv preprint arXiv:2207.12560\/} (2022).

\bibitem{goodfellow2020generative}
{\sc Goodfellow, I., Pouget-Abadie, J., Mirza, M., Xu, B., Warde-Farley, D.,
  Ozair, S., Courville, A., and Bengio, Y.}
\newblock Generative adversarial networks.
\newblock {\em Communications of the ACM 63}, 11 (2020), 139--144.

\bibitem{gretton2009covariate}
{\sc Gretton, A., Smola, A., Huang, J., Schmittfull, M., Borgwardt, K., and
  Sch{\"o}lkopf, B.}
\newblock Covariate shift by kernel mean matching.
\newblock {\em Dataset shift in machine learning 3}, 4 (2009), 5.

\bibitem{grinstein2002benchmark}
{\sc Grinstein, G.~G., Hoffman, P., Pickett, R.~M., and Laskowski, S.~J.}
\newblock Benchmark development for the evaluation of visualization for data
  mining.
\newblock {\em Information visualization in data mining and knowledge
  discovery\/} (2002), 129--176.

\bibitem{grochowski2004comparison}
{\sc Grochowski, M., and Jankowski, N.}
\newblock Comparison of instance selection algorithms ii. results and comments.
\newblock In {\em ICAISC\/} (2004).

\bibitem{gu2019using}
{\sc Gu, K., Yang, B., Ngiam, J., Le, Q., and Shlens, J.}
\newblock Using videos to evaluate image model robustness.
\newblock {\em arXiv preprint arXiv:1904.10076\/} (2019).

\bibitem{guan2021domain}
{\sc Guan, H., and Liu, M.}
\newblock Domain adaptation for medical image analysis: a survey.
\newblock {\em IEEE Transactions on Biomedical Engineering 69}, 3 (2021),
  1173--1185.

\bibitem{hamilton2020time}
{\sc Hamilton, J.~D.}
\newblock {\em Time series analysis}.
\newblock Princeton university press, 2020.

\bibitem{han2022g}
{\sc Han, X., Jiang, Z., Liu, N., and Hu, X.}
\newblock G-mixup: Graph data augmentation for graph classification.
\newblock In {\em ICML\/} (2022).

\bibitem{haviv2021bertese}
{\sc Haviv, A., Berant, J., and Globerson, A.}
\newblock Bertese: Learning to speak to bert.
\newblock In {\em EACL\/} (2021).

\bibitem{he2008adasyn}
{\sc He, H., Bai, Y., Garcia, E.~A., and Li, S.}
\newblock Adasyn: Adaptive synthetic sampling approach for imbalanced learning.
\newblock In {\em WCCI\/} (2008).

\bibitem{he2016learning}
{\sc He, Y., Tang, J., Ouyang, H., Kang, C., Yin, D., and Chang, Y.}
\newblock Learning to rewrite queries.
\newblock In {\em CIKM\/} (2016).

\bibitem{heffetz2020deepline}
{\sc Heffetz, Y., Vainshtein, R., Katz, G., and Rokach, L.}
\newblock Deepline: Automl tool for pipelines generation using deep
  reinforcement learning and hierarchical actions filtering.
\newblock In {\em KDD\/} (2020).

\bibitem{heise2014estimating}
{\sc Heise, A., Kasneci, G., and Naumann, F.}
\newblock Estimating the number and sizes of fuzzy-duplicate clusters.
\newblock In {\em CIKM\/} (2014).

\bibitem{hendrycks2019benchmarking}
{\sc Hendrycks, D., and Dietterich, T.}
\newblock Benchmarking neural network robustness to common corruptions and
  perturbations.
\newblock {\em arXiv preprint arXiv:1903.12261\/} (2019).

\bibitem{herodotou2011starfish}
{\sc Herodotou, H., Lim, H., Luo, G., Borisov, N., Dong, L., Cetin, F.~B., and
  Babu, S.}
\newblock Starfish: A self-tuning system for big data analytics.
\newblock In {\em CIDR\/} (2011).

\bibitem{ho2020denoising}
{\sc Ho, J., Jain, A., and Abbeel, P.}
\newblock Denoising diffusion probabilistic models.
\newblock In {\em NeurIPS\/} (2020).

\bibitem{ho2022cascaded}
{\sc Ho, J., Saharia, C., Chan, W., Fleet, D.~J., Norouzi, M., and Salimans,
  T.}
\newblock Cascaded diffusion models for high fidelity image generation.
\newblock {\em J. Mach. Learn. Res. 23}, 47 (2022), 1--33.

\bibitem{hooper2021cut}
{\sc Hooper, S., Wornow, M., Seah, Y.~H., Kellman, P., Xue, H., Sala, F.,
  Langlotz, C., and Re, C.}
\newblock Cut out the annotator, keep the cutout: better segmentation with weak
  supervision.
\newblock In {\em ICLR\/} (2021).

\bibitem{hsu2017unsupervised}
{\sc Hsu, W.-N., Zhang, Y., and Glass, J.}
\newblock Unsupervised domain adaptation for robust speech recognition via
  variational autoencoder-based data augmentation.
\newblock In {\em ASRU\/} (2017).

\bibitem{iwana2021empirical}
{\sc Iwana, B.~K., and Uchida, S.}
\newblock An empirical survey of data augmentation for time series
  classification with neural networks.
\newblock {\em Plos one 16}, 7 (2021), e0254841.

\bibitem{jager2021benchmark}
{\sc J{\"a}ger, S., Allhorn, A., and Bie{\ss}mann, F.}
\newblock A benchmark for data imputation methods.
\newblock {\em Frontiers in big Data 4\/} (2021), 693674.

\bibitem{jain2020overview}
{\sc Jain, A., Patel, H., Nagalapatti, L., Gupta, N., Mehta, S., Guttula, S.,
  Mujumdar, S., Afzal, S., Sharma~Mittal, R., and Munigala, V.}
\newblock Overview and importance of data quality for machine learning tasks.
\newblock In {\em KDD\/} (2020).

\bibitem{jakubik2022data}
{\sc Jakubik, J., V{\"o}ssing, M., K{\"u}hl, N., Walk, J., and Satzger, G.}
\newblock Data-centric artificial intelligence.
\newblock {\em arXiv preprint arXiv:2212.11854\/} (2022).

\bibitem{jarrahi2022principles}
{\sc Jarrahi, M.~H., Memariani, A., and Guha, S.}
\newblock The principles of data-centric ai (dcai).
\newblock {\em arXiv preprint arXiv:2211.14611\/} (2022).

\bibitem{jia2021scalability}
{\sc Jia, R., Wu, F., Sun, X., Xu, J., Dao, D., Kailkhura, B., Zhang, C., Li,
  B., and Song, D.}
\newblock Scalability vs. utility: Do we have to sacrifice one for the other in
  data importance quantification?
\newblock In {\em CVPR\/} (2021).

\bibitem{jiang2023weakly}
{\sc Jiang, M., Hou, C., Zheng, A., Hu, X., Han, S., Huang, H., He, X., Yu,
  P.~S., and Zhao, Y.}
\newblock Weakly supervised anomaly detection: A survey.
\newblock {\em arXiv preprint arXiv:2302.04549\/} (2023).

\bibitem{jiang2022fmp}
{\sc Jiang, Z., Han, X., Fan, C., Liu, Z., Zou, N., Mostafavi, A., and Hu, X.}
\newblock Fmp: Toward fair graph message passing against topology bias.
\newblock {\em arXiv preprint arXiv:2202.04187\/} (2022).

\bibitem{jiang2022generalized}
{\sc Jiang, Z., Han, X., Fan, C., Yang, F., Mostafavi, A., and Hu, X.}
\newblock Generalized demographic parity for group fairness.
\newblock In {\em ICLR\/} (2022).

\bibitem{jiang2023weight}
{\sc Jiang, Z., Han, X., Jin, H., Wang, G., Zou, N., and Hu, X.}
\newblock Weight perturbation can help fairness under distribution shift.
\newblock {\em arXiv preprint arXiv:2303.03300\/} (2023).

\bibitem{jiang2020can}
{\sc Jiang, Z., Xu, F.~F., Araki, J., and Neubig, G.}
\newblock How can we know what language models know?
\newblock {\em Transactions of the Association for Computational Linguistics
  8\/} (2020), 423--438.

\bibitem{jiang2022information}
{\sc Jiang, Z., Zhou, K., Liu, Z., Li, L., Chen, R., Choi, S.-H., and Hu, X.}
\newblock An information fusion approach to learning with instance-dependent
  label noise.
\newblock In {\em ICLR\/} (2022).

\bibitem{jumper2021highly}
{\sc Jumper, J., Evans, R., Pritzel, A., Green, T., Figurnov, M., Ronneberger,
  O., Tunyasuvunakool, K., Bates, R., {\v{Z}}{\'\i}dek, A., Potapenko, A.,
  et~al.}
\newblock Highly accurate protein structure prediction with alphafold.
\newblock {\em Nature 596}, 7873 (2021), 583--589.

\bibitem{kanamori2020dace}
{\sc Kanamori, K., Takagi, T., Kobayashi, K., and Arimura, H.}
\newblock Dace: Distribution-aware counterfactual explanation by mixed-integer
  linear optimization.
\newblock In {\em IJCAI\/} (2020).

\bibitem{kanthara2022chart}
{\sc Kanthara, S., Leong, R. T.~K., Lin, X., Masry, A., Thakkar, M., Hoque, E.,
  and Joty, S.}
\newblock Chart-to-text: A large-scale benchmark for chart summarization.
\newblock {\em arXiv preprint arXiv:2203.06486\/} (2022).

\bibitem{karimi2021algorithmic}
{\sc Karimi, A.-H., Sch{\"o}lkopf, B., and Valera, I.}
\newblock Algorithmic recourse: from counterfactual explanations to
  interventions.
\newblock In {\em FAccT\/} (2021).

\bibitem{kenton2019bert}
{\sc Kenton, J. D. M.-W.~C., and Toutanova, L.~K.}
\newblock Bert: Pre-training of deep bidirectional transformers for language
  understanding.
\newblock In {\em NAACL\/} (2019).

\bibitem{khurana2018feature}
{\sc Khurana, U., Samulowitz, H., and Turaga, D.}
\newblock Feature engineering for predictive modeling using reinforcement
  learning.
\newblock In {\em AAAI\/} (2018).

\bibitem{kim2019multiaccuracy}
{\sc Kim, M.~P., Ghorbani, A., and Zou, J.}
\newblock Multiaccuracy: Black-box post-processing for fairness in
  classification.
\newblock In {\em AIES\/} (2019).

\bibitem{kingma2021variational}
{\sc Kingma, D., Salimans, T., Poole, B., and Ho, J.}
\newblock Variational diffusion models.
\newblock In {\em NeurIPS\/} (2021).

\bibitem{koh2021wilds}
{\sc Koh, P.~W., Sagawa, S., Marklund, H., Xie, S.~M., Zhang, M., Balsubramani,
  A., Hu, W., Yasunaga, M., Phillips, R.~L., Gao, I., et~al.}
\newblock Wilds: A benchmark of in-the-wild distribution shifts.
\newblock In {\em ICML\/} (2021).

\bibitem{krishnan2019alphaclean}
{\sc Krishnan, S., and Wu, E.}
\newblock Alphaclean: Automatic generation of data cleaning pipelines.
\newblock {\em arXiv preprint arXiv:1904.11827\/} (2019).

\bibitem{krizhevsky2017imagenet}
{\sc Krizhevsky, A., Sutskever, I., and Hinton, G.~E.}
\newblock Imagenet classification with deep convolutional neural networks.
\newblock {\em Communications of the ACM 60}, 6 (2017), 84--90.

\bibitem{kumar2016join}
{\sc Kumar, A., Naughton, J., Patel, J.~M., and Zhu, X.}
\newblock To join or not to join? thinking twice about joins before feature
  selection.
\newblock In {\em SIGMOD\/} (2016).

\bibitem{kurakin2018adversarial}
{\sc Kurakin, A., Goodfellow, I.~J., and Bengio, S.}
\newblock Adversarial examples in the physical world.
\newblock In {\em Artificial intelligence safety and security}. Chapman and
  Hall/CRC, 2018, pp.~99--112.

\bibitem{kutlu2020annotator}
{\sc Kutlu, M., McDonnell, T., Elsayed, T., and Lease, M.}
\newblock Annotator rationales for labeling tasks in crowdsourcing.
\newblock {\em Journal of Artificial Intelligence Research 69\/} (2020),
  143--189.

\bibitem{lai2020dual}
{\sc Lai, K.-H., Zha, D., Li, Y., and Hu, X.}
\newblock Dual policy distillation.
\newblock In {\em IJCAI\/} (2020).

\bibitem{lai2021tods}
{\sc Lai, K.-H., Zha, D., Wang, G., Xu, J., Zhao, Y., Kumar, D., Chen, Y.,
  Zumkhawaka, P., Wan, M., Martinez, D., et~al.}
\newblock Tods: An automated time series outlier detection system.
\newblock In {\em AAAI\/} (2021).

\bibitem{lai2021revisiting}
{\sc Lai, K.-H., Zha, D., Xu, J., Zhao, Y., Wang, G., and Hu, X.}
\newblock Revisiting time series outlier detection: Definitions and benchmarks.
\newblock In {\em NeurIPS\/} (2021).

\bibitem{lai2020policy}
{\sc Lai, K.-H., Zha, D., Zhou, K., and Hu, X.}
\newblock Policy-gnn: Aggregation optimization for graph neural networks.
\newblock In {\em KDD\/} (2020).

\bibitem{lakshminarayan1996imputation}
{\sc Lakshminarayan, K., Harp, S.~A., Goldman, R.~P., Samad, T., et~al.}
\newblock Imputation of missing data using machine learning techniques.
\newblock In {\em KDD\/} (1996).

\bibitem{laugel2018comparison}
{\sc Laugel, T., Lesot, M.-J., Marsala, C., Renard, X., and Detyniecki, M.}
\newblock Comparison-based inverse classification for interpretability in
  machine learning.
\newblock In {\em IPMU\/} (2018).

\bibitem{lenzerini2002data}
{\sc Lenzerini, M.}
\newblock Data integration: A theoretical perspective.
\newblock In {\em PODS\/} (2002).

\bibitem{li2017feature}
{\sc Li, J., Cheng, K., Wang, S., Morstatter, F., Trevino, R.~P., Tang, J., and
  Liu, H.}
\newblock Feature selection: A data perspective.
\newblock {\em ACM computing surveys (CSUR) 50}, 6 (2017), 1--45.

\bibitem{li2019cleanml}
{\sc Li, P., Rao, X., Blase, J., Zhang, Y., Chu, X., and Zhang, C.}
\newblock Cleanml: A benchmark for joint data cleaning and machine learning
  [experiments and analysis].
\newblock {\em arXiv preprint arXiv:1904.09483\/} (2019), 75.

\bibitem{li2022tts}
{\sc Li, X., Metsis, V., Wang, H., and Ngu, A. H.~H.}
\newblock Tts-gan: A transformer-based time-series generative adversarial
  network.
\newblock In {\em AIME\/} (2022).

\bibitem{li2022towards}
{\sc Li, Y., Chen, Z., Zha, D., Du, M., Ni, J., Zhang, D., Chen, H., and Hu,
  X.}
\newblock Towards learning disentangled representations for time series.
\newblock In {\em KDD\/} (2022).

\bibitem{li2021automated}
{\sc Li, Y., Chen, Z., Zha, D., Zhou, K., Jin, H., Chen, H., and Hu, X.}
\newblock Automated anomaly detection via curiosity-guided search and
  self-imitation learning.
\newblock {\em IEEE Transactions on Neural Networks and Learning Systems 33}, 6
  (2021), 2365--2377.

\bibitem{li2021autood}
{\sc Li, Y., Chen, Z., Zha, D., Zhou, K., Jin, H., Chen, H., and Hu, X.}
\newblock Autood: Neural architecture search for outlier detection.
\newblock In {\em ICDE\/} (2021).

\bibitem{li2020pyodds}
{\sc Li, Y., Zha, D., Venugopal, P., Zou, N., and Hu, X.}
\newblock Pyodds: An end-to-end outlier detection system with automated machine
  learning.
\newblock In {\em WWW\/} (2020).

\bibitem{lipton2018detecting}
{\sc Lipton, Z., Wang, Y.-X., and Smola, A.}
\newblock Detecting and correcting for label shift with black box predictors.
\newblock In {\em ICML\/} (2018).

\bibitem{liu2023pre}
{\sc Liu, P., Yuan, W., Fu, J., Jiang, Z., Hayashi, H., and Neubig, G.}
\newblock Pre-train, prompt, and predict: A systematic survey of prompting
  methods in natural language processing.
\newblock {\em ACM Computing Surveys 55}, 9 (2023), 1--35.

\bibitem{liu2022rsc}
{\sc Liu, Z., Chen, S., Zhou, K., Zha, D., Huang, X., and Hu, X.}
\newblock Rsc: Accelerating graph neural networks training via randomized
  sparse computations.
\newblock {\em arXiv preprint arXiv:2210.10737\/} (2022).

\bibitem{liu2020mesa}
{\sc Liu, Z., Wei, P., Jiang, J., Cao, W., Bian, J., and Chang, Y.}
\newblock Mesa: boost ensemble imbalanced learning with meta-sampler.
\newblock In {\em NeurIPS\/} (2020).

\bibitem{lucic2022focus}
{\sc Lucic, A., Oosterhuis, H., Haned, H., and de~Rijke, M.}
\newblock Focus: Flexible optimizable counterfactual explanations for tree
  ensembles.
\newblock In {\em AAAI\/} (2022).

\bibitem{luo2018deepeye}
{\sc Luo, Y., Qin, X., Tang, N., and Li, G.}
\newblock Deepeye: Towards automatic data visualization.
\newblock In {\em 2018 IEEE 34th international conference on data engineering
  (ICDE)\/} (2018), IEEE, pp.~101--112.

\bibitem{madry2017towards}
{\sc Madry, A., Makelov, A., Schmidt, L., Tsipras, D., and Vladu, A.}
\newblock Towards deep learning models resistant to adversarial attacks.
\newblock {\em arXiv preprint arXiv:1706.06083\/} (2017).

\bibitem{yarn}
{\sc Management, C.~P.}
\newblock Clouderayarntuning.
\newblock {\em
  https://docs.cloudera.com/documentation/enterprise/latest/topics/cdh\_ig\_yarn\_tuning.html\/}
  (2023).

\bibitem{marcus2020benchmarking}
{\sc Marcus, R., Kipf, A., van Renen, A., Stoian, M., Misra, S., Kemper, A.,
  Neumann, T., and Kraska, T.}
\newblock Benchmarking learned indexes.
\newblock In {\em VLDB\/} (2020).

\bibitem{martinez2023towards}
{\sc Martinex, D., Zha, D., Tan, Q., and Hu, X.}
\newblock Towards personalized preprocessing pipeline search.
\newblock {\em arXiv preprint arXiv:2302.14329\/} (2023).

\bibitem{mazumder2022dataperf}
{\sc Mazumder, M., Banbury, C., Yao, X., Karla{\v{s}}, B., Rojas, W.~G.,
  Diamos, S., Diamos, G., He, L., Kiela, D., Jurado, D., et~al.}
\newblock Dataperf: Benchmarks for data-centric ai development.
\newblock {\em arXiv preprint arXiv:2207.10062\/} (2022).

\bibitem{meduri2020comprehensive}
{\sc Meduri, V.~V., Popa, L., Sen, P., and Sarwat, M.}
\newblock A comprehensive benchmark framework for active learning methods in
  entity matching.
\newblock In {\em SIGMOD\/} (2020).

\bibitem{mehrabi2021survey}
{\sc Mehrabi, N., Morstatter, F., Saxena, N., Lerman, K., and Galstyan, A.}
\newblock A survey on bias and fairness in machine learning.
\newblock {\em ACM Computing Surveys (CSUR) 54}, 6 (2021), 1--35.

\bibitem{meng2022interpretability}
{\sc Meng, C., Trinh, L., Xu, N., Enouen, J., and Liu, Y.}
\newblock Interpretability and fairness evaluation of deep learning models on
  mimic-iv dataset.
\newblock {\em Scientific Reports 12}, 1 (2022), 7166.

\bibitem{milutinovic2020evaluation}
{\sc Milutinovic, M., Schoenfeld, B., Martinez-Garcia, D., Ray, S., Shah, S.,
  and Yan, D.}
\newblock On evaluation of automl systems.
\newblock In {\em ICML Workshop\/} (2020).

\bibitem{mintz2009distant}
{\sc Mintz, M., Bills, S., Snow, R., and Jurafsky, D.}
\newblock Distant supervision for relation extraction without labeled data.
\newblock In {\em ACL\/} (2009).

\bibitem{miotto2018deep}
{\sc Miotto, R., Wang, F., Wang, S., Jiang, X., and Dudley, J.~T.}
\newblock Deep learning for healthcare: review, opportunities and challenges.
\newblock {\em Briefings in bioinformatics 19}, 6 (2018), 1236--1246.

\bibitem{miranda2021towards}
{\sc Miranda, L.~J.}
\newblock Towards data-centric machine learning: a short review.
\newblock {\em ljvmiranda921.github.io\/} (2021).

\bibitem{mirdita2017uniclust}
{\sc Mirdita, M., Von Den~Driesch, L., Galiez, C., Martin, M.~J., S{\"o}ding,
  J., and Steinegger, M.}
\newblock Uniclust databases of clustered and deeply annotated protein
  sequences and alignments.
\newblock {\em Nucleic acids research 45}, D1 (2017), D170--D176.

\bibitem{mnih2013playing}
{\sc Mnih, V., Kavukcuoglu, K., Silver, D., Graves, A., Antonoglou, I.,
  Wierstra, D., and Riedmiller, M.}
\newblock Playing atari with deep reinforcement learning.
\newblock {\em arXiv preprint arXiv:1312.5602\/} (2013).

\bibitem{moosavi2016deepfool}
{\sc Moosavi-Dezfooli, S.-M., Fawzi, A., and Frossard, P.}
\newblock Deepfool: a simple and accurate method to fool deep neural networks.
\newblock In {\em CVPR\/} (2016).

\bibitem{nanni2021comparison}
{\sc Nanni, L., Paci, M., Brahnam, S., and Lumini, A.}
\newblock Comparison of different image data augmentation approaches.
\newblock {\em Journal of imaging 7}, 12 (2021), 254.

\bibitem{nargesian2018table}
{\sc Nargesian, F., Zhu, E., Pu, K.~Q., and Miller, R.~J.}
\newblock Table union search on open data.
\newblock In {\em VLDB\/} (2018).

\bibitem{datacentricaihub}
{\sc Ng, A.}
\newblock Data-centric ai resource hub.
\newblock {\em Snorkel AI. Available online: https://snorkel.ai/(accessed on 8
  February 2023)\/} (2021).

\bibitem{landingai}
{\sc Ng, A.}
\newblock Landing ai.
\newblock {\em Landing AI. Available online: https://landing.ai/(accessed on 8
  February 2023)\/} (2023).

\bibitem{ng2021data}
{\sc Ng, A., Laird, D., and He, L.}
\newblock Data-centric ai competition.
\newblock {\em DeepLearning AI. Available online:
  https://https-deeplearning-ai. github. io/data-centric-comp/(accessed on 8
  December 2021)\/} (2021).

\bibitem{obukhov2020quality}
{\sc Obukhov, A., and Krasnyanskiy, M.}
\newblock Quality assessment method for gan based on modified metrics inception
  score and fr{\'e}chet inception distance.
\newblock In {\em CoMeSySo\/} (2020).

\bibitem{gpt4}
{\sc OpenAI}.
\newblock Gpt-4 technical report, 2023.

\bibitem{otles2021mind}
{\sc Otles, E., Oh, J., Li, B., Bochinski, M., Joo, H., Ortwine, J., Shenoy,
  E., Washer, L., Young, V.~B., Rao, K., et~al.}
\newblock Mind the performance gap: examining dataset shift during prospective
  validation.
\newblock In {\em MLHC\/} (2021).

\bibitem{ouyang2022training}
{\sc Ouyang, L., Wu, J., Jiang, X., Almeida, D., Wainwright, C.~L., Mishkin,
  P., Zhang, C., Agarwal, S., Slama, K., Ray, A., et~al.}
\newblock Training language models to follow instructions with human feedback.
\newblock In {\em NeurIPS\/} (2022).

\bibitem{ozbayoglu2020deep}
{\sc Ozbayoglu, A.~M., Gudelek, M.~U., and Sezer, O.~B.}
\newblock Deep learning for financial applications: A survey.
\newblock {\em Applied Soft Computing 93\/} (2020), 106384.

\bibitem{pang2021deep}
{\sc Pang, G., Shen, C., Cao, L., and Hengel, A. V.~D.}
\newblock Deep learning for anomaly detection: A review.
\newblock {\em ACM computing surveys (CSUR) 54}, 2 (2021), 1--38.

\bibitem{papernot2017practical}
{\sc Papernot, N., McDaniel, P., Goodfellow, I., Jha, S., Celik, Z.~B., and
  Swami, A.}
\newblock Practical black-box attacks against machine learning.
\newblock In {\em ASIACCS\/} (2017).

\bibitem{pawelczyk2021carla}
{\sc Pawelczyk, M., Bielawski, S., Heuvel, J. v.~d., Richter, T., and Kasneci,
  G.}
\newblock Carla: a python library to benchmark algorithmic recourse and
  counterfactual explanation algorithms.
\newblock {\em arXiv preprint arXiv:2108.00783\/} (2021).

\bibitem{pedrozo2018adaptive}
{\sc Pedrozo, W.~G., Nievola, J.~C., and Ribeiro, D.~C.}
\newblock An adaptive approach for index tuning with learning classifier
  systems on hybrid storage environments.
\newblock In {\em HAIS\/} (2018).

\bibitem{pinkel2015rodi}
{\sc Pinkel, C., Binnig, C., Jim{\'e}nez-Ruiz, E., May, W., Ritze, D.,
  Skj{\ae}veland, M.~G., Solimando, A., and Kharlamov, E.}
\newblock Rodi: A benchmark for automatic mapping generation in
  relational-to-ontology data integration.
\newblock In {\em ESWC\/} (2015).

\bibitem{pipino2002data}
{\sc Pipino, L.~L., Lee, Y.~W., and Wang, R.~Y.}
\newblock Data quality assessment.
\newblock {\em Communications of the ACM 45}, 4 (2002), 211--218.

\bibitem{poess2014tpc}
{\sc Poess, M., Rabl, T., Jacobsen, H.-A., and Caufield, B.}
\newblock Tpc-di: the first industry benchmark for data integration.
\newblock In {\em VLDB\/} (2014).

\bibitem{polyzotis2021can}
{\sc Polyzotis, N., and Zaharia, M.}
\newblock What can data-centric ai learn from data and ml engineering?
\newblock {\em arXiv preprint arXiv:2112.06439\/} (2021).

\bibitem{poyiadzi2020face}
{\sc Poyiadzi, R., Sokol, K., Santos-Rodriguez, R., De~Bie, T., and Flach, P.}
\newblock Face: feasible and actionable counterfactual explanations.
\newblock In {\em AAAI\/} (2020).

\bibitem{press_2022}
{\sc Press, G.}
\newblock Cleaning big data: Most time-consuming, least enjoyable data science
  task, survey says, Oct 2022.

\bibitem{prusa2015using}
{\sc Prusa, J., Khoshgoftaar, T.~M., Dittman, D.~J., and Napolitano, A.}
\newblock Using random undersampling to alleviate class imbalance on tweet
  sentiment data.
\newblock In {\em IRI\/} (2015).

\bibitem{radford2018improving}
{\sc Radford, A., Narasimhan, K., Salimans, T., Sutskever, I., et~al.}
\newblock Improving language understanding by generative pre-training.
\newblock {\em OpenAI\/} (2018).

\bibitem{radford2019language}
{\sc Radford, A., Wu, J., Child, R., Luan, D., Amodei, D., Sutskever, I.,
  et~al.}
\newblock Language models are unsupervised multitask learners.
\newblock {\em OpenAI\/} (2019).

\bibitem{snorkelai}
{\sc Ratner, A.}
\newblock Scale ai.
\newblock {\em Snorkel AI. Available online: https://snorkel.ai/(accessed on 8
  February 2023)\/} (2023).

\bibitem{ratner2017snorkel}
{\sc Ratner, A., Bach, S.~H., Ehrenberg, H., Fries, J., Wu, S., and R{\'e}, C.}
\newblock Snorkel: Rapid training data creation with weak supervision.
\newblock In {\em VLDB\/} (2017).

\bibitem{ratner2016data}
{\sc Ratner, A.~J., De~Sa, C.~M., Wu, S., Selsam, D., and R{\'e}, C.}
\newblock Data programming: Creating large training sets, quickly.
\newblock {\em NeurIPS\/} (2016).

\bibitem{ren2021survey}
{\sc Ren, P., Xiao, Y., Chang, X., Huang, P.-Y., Li, Z., Gupta, B.~B., Chen,
  X., and Wang, X.}
\newblock A survey of deep active learning.
\newblock {\em ACM computing surveys (CSUR) 54}, 9 (2021), 1--40.

\bibitem{riquelme2003finding}
{\sc Riquelme, J.~C., Aguilar-Ruiz, J.~S., and Toro, M.}
\newblock Finding representative patterns with ordered projections.
\newblock {\em pattern recognition 36}, 4 (2003), 1009--1018.

\bibitem{rombach2022high}
{\sc Rombach, R., Blattmann, A., Lorenz, D., Esser, P., and Ommer, B.}
\newblock High-resolution image synthesis with latent diffusion models.
\newblock In {\em CVPR\/} (2022).

\bibitem{sadiq2018data}
{\sc Sadiq, S., Dasu, T., Dong, X.~L., Freire, J., Ilyas, I.~F., Link, S.,
  Miller, M.~J., Naumann, F., Zhou, X., and Srivastava, D.}
\newblock Data quality: The role of empiricism.
\newblock {\em ACM SIGMOD Record 46}, 4 (2018), 35--43.

\bibitem{sadri2020online}
{\sc Sadri, Z., Gruenwald, L., and Leal, E.}
\newblock Online index selection using deep reinforcement learning for a
  cluster database.
\newblock In {\em ICDE Workshop\/} (2020).

\bibitem{saenko2010adapting}
{\sc Saenko, K., Kulis, B., Fritz, M., and Darrell, T.}
\newblock Adapting visual category models to new domains.
\newblock In {\em ECCV\/} (2010).

\bibitem{sagadeeva2021sliceline}
{\sc Sagadeeva, S., and Boehm, M.}
\newblock Sliceline: Fast, linear-algebra-based slice finding for ml model
  debugging.
\newblock In {\em SIGMOD\/} (2021).

\bibitem{salau2019feature}
{\sc Salau, A.~O., and Jain, S.}
\newblock Feature extraction: a survey of the types, techniques, applications.
\newblock In {\em ICSC\/} (2019).

\bibitem{sambasivan2021everyone}
{\sc Sambasivan, N., Kapania, S., Highfill, H., Akrong, D., Paritosh, P., and
  Aroyo, L.~M.}
\newblock “everyone wants to do the model work, not the data work”: Data
  cascades in high-stakes ai.
\newblock In {\em CHI\/} (2021).

\bibitem{sangkloy2017scribbler}
{\sc Sangkloy, P., Lu, J., Fang, C., Yu, F., and Hays, J.}
\newblock Scribbler: Controlling deep image synthesis with sketch and color.
\newblock In {\em CVPR\/} (2017).

\bibitem{santelices2013quantitative}
{\sc Santelices, R., Zhang, Y., Jiang, S., Cai, H., and Zhang, Y.-j.}
\newblock Quantitative program slicing: Separating statements by relevance.
\newblock In {\em ICSE\/} (2013).

\bibitem{saporta2002data}
{\sc Saporta, G.}
\newblock Data fusion and data grafting.
\newblock {\em Computational statistics \& data analysis 38}, 4 (2002),
  465--473.

\bibitem{schelter2018automating}
{\sc Schelter, S., Lange, D., Schmidt, P., Celikel, M., Biessmann, F., and
  Grafberger, A.}
\newblock Automating large-scale data quality verification.
\newblock In {\em VLDB\/} (2018).

\bibitem{schick2020exploiting}
{\sc Schick, T., and Sch{\"u}tze, H.}
\newblock Exploiting cloze questions for few shot text classification and
  natural language inference.
\newblock {\em arXiv preprint arXiv:2001.07676\/} (2020).

\bibitem{schick2020few}
{\sc Schick, T., and Sch{\"u}tze, H.}
\newblock Few-shot text generation with pattern-exploiting training.
\newblock {\em arXiv preprint arXiv:2012.11926\/} (2020).

\bibitem{schick2020s}
{\sc Schick, T., and Sch{\"u}tze, H.}
\newblock It's not just size that matters: Small language models are also
  few-shot learners.
\newblock {\em arXiv preprint arXiv:2009.07118\/} (2020).

\bibitem{schnapp2021active}
{\sc Schnapp, S., and Sabato, S.}
\newblock Active feature selection for the mutual information criterion.
\newblock In {\em AAAI\/} (2021).

\bibitem{seedat2022dc}
{\sc Seedat, N., Imrie, F., and van~der Schaar, M.}
\newblock Dc-check: A data-centric ai checklist to guide the development of
  reliable machine learning systems.
\newblock {\em arXiv preprint arXiv:2211.05764\/} (2022).

\bibitem{shafahi2018poison}
{\sc Shafahi, A., Huang, W.~R., Najibi, M., Suciu, O., Studer, C., Dumitras,
  T., and Goldstein, T.}
\newblock Poison frogs! targeted clean-label poisoning attacks on neural
  networks.
\newblock In {\em NeurIPS\/} (2018).

\bibitem{shankar2021image}
{\sc Shankar, V., Dave, A., Roelofs, R., Ramanan, D., Recht, B., and Schmidt,
  L.}
\newblock Do image classifiers generalize across time?
\newblock In {\em ICCV\/} (2021).

\bibitem{sharma2019certifai}
{\sc Sharma, S., Henderson, J., and Ghosh, J.}
\newblock Certifai: Counterfactual explanations for robustness, transparency,
  interpretability, and fairness of artificial intelligence models.
\newblock {\em arXiv preprint arXiv:1905.07857\/} (2019).

\bibitem{shen2021towardsnli}
{\sc Shen, L., Shen, E., Luo, Y., Yang, X., Hu, X., Zhang, X., Tai, Z., and
  Wang, J.}
\newblock Towards natural language interfaces for data visualization: A survey.
\newblock {\em arXiv preprint arXiv:2109.03506\/} (2021).

\bibitem{shen2021towards}
{\sc Shen, Z., Liu, J., He, Y., Zhang, X., Xu, R., Yu, H., and Cui, P.}
\newblock Towards out-of-distribution generalization: A survey.
\newblock {\em arXiv preprint arXiv:2108.13624\/} (2021).

\bibitem{shorten2019survey}
{\sc Shorten, C., and Khoshgoftaar, T.~M.}
\newblock A survey on image data augmentation for deep learning.
\newblock {\em Journal of big data 6}, 1 (2019), 1--48.

\bibitem{shorten2021text}
{\sc Shorten, C., Khoshgoftaar, T.~M., and Furht, B.}
\newblock Text data augmentation for deep learning.
\newblock {\em Journal of big Data 8\/} (2021), 1--34.

\bibitem{sohoni2020no}
{\sc Sohoni, N., Dunnmon, J., Angus, G., Gu, A., and R{\'e}, C.}
\newblock No subclass left behind: Fine-grained robustness in coarse-grained
  classification problems.
\newblock In {\em NeurIPS\/} (2020).

\bibitem{souza2019data}
{\sc Souza, J. T.~d., Francisco, A. C.~d., Piekarski, C.~M., and Prado, G.
  F.~d.}
\newblock Data mining and machine learning to promote smart cities: A
  systematic review from 2000 to 2018.
\newblock {\em Sustainability 11}, 4 (2019), 1077.

\bibitem{srinivasan2021snowy}
{\sc Srinivasan, A., and Setlur, V.}
\newblock Snowy: Recommending utterances for conversational visual analysis.
\newblock In {\em SIGCHI\/} (2021).

\bibitem{srivastava2022beyond}
{\sc Srivastava, A., Rastogi, A., Rao, A., Shoeb, A. A.~M., Abid, A., Fisch,
  A., Brown, A.~R., Santoro, A., Gupta, A., Garriga-Alonso, A., et~al.}
\newblock Beyond the imitation game: Quantifying and extrapolating the
  capabilities of language models.
\newblock {\em arXiv preprint arXiv:2206.04615\/} (2022).

\bibitem{stonebraker2013data}
{\sc Stonebraker, M., Bruckner, D., Ilyas, I.~F., Beskales, G., Cherniack, M.,
  Zdonik, S.~B., Pagan, A., and Xu, S.}
\newblock Data curation at scale: the data tamer system.
\newblock In {\em CIDR\/} (2013).

\bibitem{stonebraker2018data}
{\sc Stonebraker, M., Ilyas, I.~F., et~al.}
\newblock Data integration: The current status and the way forward.
\newblock {\em IEEE Data Eng. Bull. 41}, 2 (2018), 3--9.

\bibitem{sugiyama2007covariate}
{\sc Sugiyama, M., Krauledat, M., and M{\"u}ller, K.-R.}
\newblock Covariate shift adaptation by importance weighted cross validation.
\newblock {\em Journal of Machine Learning Research 8}, 5 (2007).

\bibitem{sun2019end}
{\sc Sun, J., and Li, G.}
\newblock An end-to-end learning-based cost estimator.
\newblock In {\em VLDB\/} (2019).

\bibitem{sutton2012introduction}
{\sc Sutton, O.}
\newblock Introduction to k nearest neighbour classification and condensed
  nearest neighbour data reduction.
\newblock {\em University lectures, University of Leicester 1\/} (2012).

\bibitem{tan2023bring}
{\sc Tan, Q., Zhang, X., Liu, N., Zha, D., Li, L., Chen, R., Choi, S.-H., and
  Hu, X.}
\newblock Bring your own view: Graph neural networks for link prediction with
  personalized subgraph selection.
\newblock In {\em WSDM\/} (2023).

\bibitem{tang2011semi}
{\sc Tang, W., and Lease, M.}
\newblock Semi-supervised consensus labeling for crowdsourcing.
\newblock In {\em SIGIR Workshop\/} (2011).

\bibitem{tao2021benchmarking}
{\sc Tao, Y., McKenna, R., Hay, M., Machanavajjhala, A., and Miklau, G.}
\newblock Benchmarking differentially private synthetic data generation
  algorithms.
\newblock {\em arXiv preprint arXiv:2112.09238\/} (2021).

\bibitem{thaseen2017intrusion}
{\sc Thaseen, I.~S., and Kumar, C.~A.}
\newblock Intrusion detection model using fusion of chi-square feature
  selection and multi class svm.
\newblock {\em Journal of King Saud University-Computer and Information
  Sciences 29}, 4 (2017), 462--472.

\bibitem{thirumuruganathan2020data}
{\sc Thirumuruganathan, S., Tang, N., Ouzzani, M., and Doan, A.}
\newblock Data curation with deep learning.
\newblock In {\em EDBT\/} (2020).

\bibitem{thusoo2010data}
{\sc Thusoo, A., Shao, Z., Anthony, S., Borthakur, D., Jain, N., Sen~Sarma, J.,
  Murthy, R., and Liu, H.}
\newblock Data warehousing and analytics infrastructure at facebook.
\newblock In {\em SIGMOD\/} (2010).

\bibitem{valentin2000db2}
{\sc Valentin, G., Zuliani, M., Zilio, D.~C., Lohman, G., and Skelley, A.}
\newblock Db2 advisor: An optimizer smart enough to recommend its own indexes.
\newblock In {\em ICDE\/} (2000).

\bibitem{van2017automatic}
{\sc Van~Aken, D., Pavlo, A., Gordon, G.~J., and Zhang, B.}
\newblock Automatic database management system tuning through large-scale
  machine learning.
\newblock In {\em SIGMOD\/} (2017).

\bibitem{varia2014overview}
{\sc Varia, J., Mathew, S., et~al.}
\newblock Overview of amazon web services.
\newblock {\em Amazon Web Services\/} (2014).

\bibitem{vijayan2022blood}
{\sc Vijayan, A., Fatima, S., Sowmya, A., and Vafaee, F.}
\newblock Blood-based transcriptomic signature panel identification for cancer
  diagnosis: benchmarking of feature extraction methods.
\newblock {\em Briefings in Bioinformatics 23}, 5 (2022), bbac315.

\bibitem{voulodimos2018deep}
{\sc Voulodimos, A., Doulamis, N., Doulamis, A., Protopapadakis, E., et~al.}
\newblock Deep learning for computer vision: A brief review.
\newblock {\em Computational intelligence and neuroscience 2018\/} (2018).

\bibitem{wachter2017counterfactual}
{\sc Wachter, S., Mittelstadt, B., and Russell, C.}
\newblock Counterfactual explanations without opening the black box: Automated
  decisions and the gdpr.
\newblock {\em Harv. JL \& Tech. 31\/} (2017), 841.

\bibitem{waldner2019comparison}
{\sc Waldner, M., Diehl, A., Gra{\v{c}}anin, D., Splechtna, R., Delrieux, C.,
  and Matkovi{\'c}, K.}
\newblock A comparison of radial and linear charts for visualizing daily
  patterns.
\newblock {\em IEEE transactions on visualization and computer graphics 26}, 1
  (2019).

\bibitem{wallace2019universal}
{\sc Wallace, E., Feng, S., Kandpal, N., Gardner, M., and Singh, S.}
\newblock Universal adversarial triggers for attacking and analyzing nlp.
\newblock In {\em IJCNLP\/} (2019).

\bibitem{wan2022processing}
{\sc Wan, M., Zha, D., Liu, N., and Zou, N.}
\newblock In-processing modeling techniques for machine learning fairness: A
  survey.
\newblock {\em ACM Transactions on Knowledge Discovery from Data (TKDD)\/}
  (2022).

\bibitem{scaleai}
{\sc Wang, A.}
\newblock Scale ai.
\newblock {\em Scale AI. Available online: https://scale.com/(accessed on 8
  February 2023)\/} (2023).

\bibitem{wang2022bed}
{\sc Wang, G., Bhat, Z.~P., Jiang, Z., Chen, Y.-W., Zha, D., Reyes, A.~C.,
  Niktash, A., Ulkar, G., Okman, E., Cai, X., et~al.}
\newblock Bed: A real-time object detection system for edge devices.
\newblock In {\em CIKM\/} (2022), pp.~4994--4998.

\bibitem{wang2022accelerating}
{\sc Wang, G., Chuang, Y.-N., Du, M., Yang, F., Zhou, Q., Tripathi, P., Cai,
  X., and Hu, X.}
\newblock Accelerating shapley explanation via contributive cooperator
  selection.
\newblock In {\em ICML\/} (2022).

\bibitem{wang2012crowder}
{\sc Wang, J., Kraska, T., Franklin, M.~J., and Feng, J.}
\newblock Crowder: crowdsourcing entity resolution.
\newblock In {\em VLDB\/} (2012).

\bibitem{wang2015embedded}
{\sc Wang, S., Tang, J., and Liu, H.}
\newblock Embedded unsupervised feature selection.
\newblock In {\em AAAI\/} (2015).

\bibitem{wang2022usb}
{\sc Wang, Y., Chen, H., Fan, Y., Wang, S., Tao, R., Hou, W., Wang, R., Yang,
  L., Zhou, Z., Guo, L.-Z., et~al.}
\newblock Usb: A unified semi-supervised learning benchmark for classification.
\newblock In {\em NeurIPS\/} (2022).

\bibitem{wang2021crowdsourcing}
{\sc Wang, Y., Wang, Q., Huang, H., Huang, W., Chen, Y., McGarvey, P.~B., Wu,
  C.~H., Arighi, C.~N., and Consortium, U.}
\newblock A crowdsourcing open platform for literature curation in uniprot.
\newblock {\em PLoS biology 19}, 12 (2021), e3001464.

\bibitem{wang2017time}
{\sc Wang, Z., Yan, W., and Oates, T.}
\newblock Time series classification from scratch with deep neural networks: A
  strong baseline.
\newblock In {\em IJCNN\/} (2017).

\bibitem{webb2018deep}
{\sc Webb, S., et~al.}
\newblock Deep learning for biology.
\newblock {\em Nature 554}, 7693 (2018), 555--557.

\bibitem{wen2021time}
{\sc Wen, Q., Sun, L., Yang, F., Song, X., Gao, J., Wang, X., and Xu, H.}
\newblock Time series data augmentation for deep learning: A survey.
\newblock In {\em IJCAI\/} (2021).

\bibitem{whang2023data}
{\sc Whang, S.~E., Roh, Y., Song, H., and Lee, J.-G.}
\newblock Data collection and quality challenges in deep learning: A
  data-centric ai perspective.
\newblock In {\em VLDB\/} (2023).

\bibitem{white2012hadoop}
{\sc White, T.}
\newblock {\em Hadoop: The definitive guide}.
\newblock " O'Reilly Media, Inc.", 2012.

\bibitem{winston1984artificial}
{\sc Winston, P.~H.}
\newblock {\em Artificial intelligence}.
\newblock Addison-Wesley Longman Publishing Co., Inc., 1984.

\bibitem{wongsuphasawat2015voyager}
{\sc Wongsuphasawat, K., Moritz, D., Anand, A., Mackinlay, J., Howe, B., and
  Heer, J.}
\newblock Voyager: Exploratory analysis via faceted browsing of visualization
  recommendations.
\newblock {\em IEEE transactions on visualization and computer graphics 22}, 1
  (2015), 649--658.

\bibitem{xanthopoulos2013linear}
{\sc Xanthopoulos, P., Pardalos, P.~M., Trafalis, T.~B., Xanthopoulos, P.,
  Pardalos, P.~M., and Trafalis, T.~B.}
\newblock Linear discriminant analysis.
\newblock {\em Robust data mining\/} (2013), 27--33.

\bibitem{xing2021fairness}
{\sc Xing, X., Liu, H., Chen, C., and Li, J.}
\newblock Fairness-aware unsupervised feature selection.
\newblock In {\em CIKM\/} (2021).

\bibitem{xue2022knowledge}
{\sc Xue, B., and Zou, L.}
\newblock Knowledge graph quality management: a comprehensive survey.
\newblock {\em IEEE Transactions on Knowledge and Data Engineering\/} (2022).

\bibitem{yan2015feature}
{\sc Yan, K., and Zhang, D.}
\newblock Feature selection and analysis on correlated gas sensor data with
  recursive feature elimination.
\newblock {\em Sensors and Actuators B: Chemical 212\/} (2015), 353--363.

\bibitem{yang2018benchmark}
{\sc Yang, Y., and Loog, M.}
\newblock A benchmark and comparison of active learning for logistic
  regression.
\newblock {\em Pattern Recognition 83\/} (2018), 401--415.

\bibitem{ying2019overview}
{\sc Ying, X.}
\newblock An overview of overfitting and its solutions.
\newblock {\em Journal of physics: Conference series 1168\/} (2019), 022022.

\bibitem{yoo2020rethinking}
{\sc Yoo, J., Ahn, N., and Sohn, K.-A.}
\newblock Rethinking data augmentation for image super-resolution: A
  comprehensive analysis and a new strategy.
\newblock In {\em CVPR\/} (2020).

\bibitem{yoo2020searching}
{\sc Yoo, J.~Y., Morris, J.~X., Lifland, E., and Qi, Y.}
\newblock Searching for a search method: Benchmarking search algorithms for
  generating nlp adversarial examples.
\newblock {\em arXiv preprint arXiv:2009.06368\/} (2020).

\bibitem{yoo2021gpt3mix}
{\sc Yoo, K.~M., Park, D., Kang, J., Lee, S.-W., and Park, W.}
\newblock Gpt3mix: Leveraging large-scale language models for text
  augmentation.
\newblock In {\em EMNLP\/} (2021).

\bibitem{yuan2021bartscore}
{\sc Yuan, W., Neubig, G., and Liu, P.}
\newblock Bartscore: Evaluating generated text as text generation.
\newblock In {\em NeurIPS\/} (2021).

\bibitem{yuen2011survey}
{\sc Yuen, M.-C., King, I., and Leung, K.-S.}
\newblock A survey of crowdsourcing systems.
\newblock In {\em PASSAT\/} (2011).

\bibitem{ZahariaXinEtAl16cacm}
{\sc Zaharia, M., Xin, R.~S., Wendell, P., Das, T., Armbrust, M., Dave, A.,
  Meng, X., Rosen, J., Venkataraman, S., Franklin, M.~J., Ghodsi, A., Gonzalez,
  J., Shenker, S., and Stoica, I.}
\newblock {Apache Spark}: A unified engine for big data processing.
\newblock {\em Communications of the ACM 59\/} (2016).

\bibitem{zeng1998stratal}
{\sc Zeng, H., Henry, S.~C., and Riola, J.~P.}
\newblock Stratal slicing, part ii: Real 3-d seismic data.
\newblock {\em Geophysics 63}, 2 (1998), 514--522.

\bibitem{zeng2021evaluation}
{\sc Zeng, Z., Moh, P., Du, F., Hoffswell, J., Lee, T.~Y., Malik, S., Koh, E.,
  and Battle, L.}
\newblock An evaluation-focused framework for visualization recommendation
  algorithms.
\newblock {\em IEEE Transactions on Visualization and Computer Graphics 28}, 1
  (2021), 346--356.

\bibitem{zha2023data}
{\sc Zha, D., Bhat, Z.~P., Lai, K.-H., Yang, F., and Hu, X.}
\newblock Data-centric ai: Perspectives and challenges.
\newblock {\em arXiv preprint arXiv:2301.04819\/} (2023).

\bibitem{zha2022autoshard}
{\sc Zha, D., Feng, L., Bhushanam, B., Choudhary, D., Nie, J., Tian, Y., Chae,
  J., Ma, Y., Kejariwal, A., and Hu, X.}
\newblock Autoshard: Automated embedding table sharding for recommender
  systems.
\newblock In {\em KDD\/} (2022).

\bibitem{zhadreamshard}
{\sc Zha, D., Feng, L., Tan, Q., Liu, Z., Lai, K.-H., Bhushanam, B., Tian, Y.,
  Kejariwal, A., and Hu, X.}
\newblock Dreamshard: Generalizable embedding table placement for recommender
  systems.
\newblock In {\em NeurIPS\/} (2022).

\bibitem{zha2021rlcard}
{\sc Zha, D., Lai, K.-H., Huang, S., Cao, Y., Reddy, K., Vargas, J., Nguyen,
  A., Wei, R., Guo, J., and Hu, X.}
\newblock Rlcard: a platform for reinforcement learning in card games.
\newblock In {\em IJCAI\/} (2021).

\bibitem{zha2022towards}
{\sc Zha, D., Lai, K.-H., Tan, Q., Ding, S., Zou, N., and Hu, X.~B.}
\newblock Towards automated imbalanced learning with deep hierarchical
  reinforcement learning.
\newblock In {\em CIKM\/} (2022).

\bibitem{zha2020meta}
{\sc Zha, D., Lai, K.-H., Wan, M., and Hu, X.}
\newblock Meta-aad: Active anomaly detection with deep reinforcement learning.
\newblock In {\em ICDM\/} (2020).

\bibitem{zha2019experience}
{\sc Zha, D., Lai, K.-H., Zhou, K., and Hu, X.}
\newblock Experience replay optimization.
\newblock In {\em IJCAI\/} (2019).

\bibitem{zha2021simplifying}
{\sc Zha, D., Lai, K.-H., Zhou, K., and Hu, X.}
\newblock Simplifying deep reinforcement learning via self-supervision.
\newblock {\em arXiv preprint arXiv:2106.05526\/} (2021).

\bibitem{zha2022towards2}
{\sc Zha, D., Lai, K.-H., Zhou, K., and Hu, X.}
\newblock Towards similarity-aware time-series classification.
\newblock In {\em SDM\/} (2022).

\bibitem{zha2019multi}
{\sc Zha, D., and Li, C.}
\newblock Multi-label dataless text classification with topic modeling.
\newblock {\em Knowledge and Information Systems 61\/} (2019), 137--160.

\bibitem{zha2021rank}
{\sc Zha, D., Ma, W., Yuan, L., Hu, X., and Liu, J.}
\newblock Rank the episodes: A simple approach for exploration in
  procedurally-generated environments.
\newblock In {\em ICLR\/} (2021).

\bibitem{zha2021autovideo}
{\sc Zha, D., Pervaiz~Bhat, Z., Chen, Y.-W., Wang, Y., Ding, S., Jain, A.~K.,
  Qazim~Bhat, M., Lai, K.-H., Chen, J., et~al.}
\newblock Autovideo: An automated video action recognition system.
\newblock In {\em IJCAI\/} (2022).

\bibitem{zha2021douzero}
{\sc Zha, D., Xie, J., Ma, W., Zhang, S., Lian, X., Hu, X., and Liu, J.}
\newblock Douzero: Mastering doudizhu with self-play deep reinforcement
  learning.
\newblock In {\em ICML\/} (2021).

\bibitem{zhang2018mixup}
{\sc Zhang, H., Cisse, M., Dauphin, Y.~N., and Lopez-Paz, D.}
\newblock mixup: Beyond empirical risk minimization.
\newblock In {\em ICLR\/} (2018).

\bibitem{zhang2019self}
{\sc Zhang, H., Goodfellow, I., Metaxas, D., and Odena, A.}
\newblock Self-attention generative adversarial networks.
\newblock In {\em IICML\/} (2019).

\bibitem{zhang2022survey}
{\sc Zhang, J., Hsieh, C.-Y., Yu, Y., Zhang, C., and Ratner, A.}
\newblock A survey on programmatic weak supervision.
\newblock {\em arXiv preprint arXiv:2202.05433\/} (2022).

\bibitem{zhang2019deep}
{\sc Zhang, S., Yao, L., Sun, A., and Tay, Y.}
\newblock Deep learning based recommender system: A survey and new
  perspectives.
\newblock {\em ACM computing surveys (CSUR) 52}, 1 (2019), 1--38.

\bibitem{zhang2022facilitating}
{\sc Zhang, X., Chang, Z., Li, Y., Wu, H., Tan, J., Li, F., and Cui, B.}
\newblock Facilitating database tuning with hyper-parameter optimization: a
  comprehensive experimental evaluation.
\newblock In {\em VLDB\/} (2022).

\bibitem{zhang2019active}
{\sc Zhang, X., Mei, C., Chen, D., Yang, Y., and Li, J.}
\newblock Active incremental feature selection using a fuzzy-rough-set-based
  information entropy.
\newblock {\em IEEE Transactions on Fuzzy Systems 28}, 5 (2019), 901--915.

\bibitem{zhang2015character}
{\sc Zhang, X., Zhao, J., and LeCun, Y.}
\newblock Character-level convolutional networks for text classification.
\newblock In {\em NeurIPS\/} (2015).

\bibitem{zhang2016missing}
{\sc Zhang, Z.}
\newblock Missing data imputation: focusing on single imputation.
\newblock {\em Annals of translational medicine 4}, 1 (2016).

\bibitem{zhou2020graph}
{\sc Zhou, J., Cui, G., Hu, S., Zhang, Z., Yang, C., Liu, Z., Wang, L., Li, C.,
  and Sun, M.}
\newblock Graph neural networks: A review of methods and applications.
\newblock {\em AI open 1\/} (2020), 57--81.

\bibitem{zhou2020towards}
{\sc Zhou, K., Huang, X., Li, Y., Zha, D., Chen, R., and Hu, X.}
\newblock Towards deeper graph neural networks with differentiable group
  normalization.
\newblock In {\em NeurIPS\/} (2020).

\bibitem{zhou2021dirichlet}
{\sc Zhou, K., Huang, X., Zha, D., Chen, R., Li, L., Choi, S.-H., and Hu, X.}
\newblock Dirichlet energy constrained learning for deep graph neural networks.
\newblock In {\em NeurIPS\/} (2021).

\bibitem{zhou2021multi}
{\sc Zhou, K., Song, Q., Huang, X., Zha, D., Zou, N., and Hu, X.}
\newblock Multi-channel graph neural networks.
\newblock In {\em IJCAI\/} (2021).

\bibitem{zhou2021dbmind}
{\sc Zhou, X., Jin, L., Sun, J., Zhao, X., Yu, X., Feng, J., Li, S., Wang, T.,
  Li, K., and Liu, L.}
\newblock Dbmind: A self-driving platform in opengauss.
\newblock In {\em VLDB\/} (2021).

\bibitem{zhou2004democratic}
{\sc Zhou, Y., and Goldman, S.}
\newblock Democratic co-learning.
\newblock In {\em ICTAI\/} (2004).

\bibitem{zhu2015aligning}
{\sc Zhu, Y., Kiros, R., Zemel, R., Salakhutdinov, R., Urtasun, R., Torralba,
  A., and Fidler, S.}
\newblock Aligning books and movies: Towards story-like visual explanations by
  watching movies and reading books.
\newblock In {\em CVPR\/} (2015).

\bibitem{zoller2021benchmark}
{\sc Z{\"o}ller, M.-A., and Huber, M.~F.}
\newblock Benchmark and survey of automated machine learning frameworks.
\newblock {\em Journal of artificial intelligence research 70\/} (2021),
  409--472.

\bibitem{zoph2020rethinking}
{\sc Zoph, B., Ghiasi, G., Lin, T.-Y., Cui, Y., Liu, H., Cubuk, E.~D., and Le,
  Q.}
\newblock Rethinking pre-training and self-training.
\newblock In {\em NeurIPS\/} (2020).

\end{thebibliography}
